\documentclass[a4paper, 12pt]{article}
\usepackage[a4paper,top=2.5cm,bottom=2.5cm,left=2cm,right=2cm,marginparwidth=2cm]{geometry}
\usepackage[english]{babel}
\usepackage[utf8x]{inputenc}
\usepackage{listings}
\usepackage{enumitem}
\usepackage{graphicx,wrapfig}
\usepackage{float}
\usepackage{amsmath}
\usepackage[colorinlistoftodos]{todonotes}
\usepackage[hyphens]{url}
\usepackage[colorlinks=true, allcolors=blue]{hyperref}
\usepackage{listings}
\usepackage{graphicx}
\hypersetup{breaklinks=true}
\usepackage{cite}
\usepackage{caption}
\usepackage{subcaption}

\title{Deployment of ML Models using Kubeflow on Different Cloud Providers}
\author{
Aditya Pandey, Maitreya Sonawane, Sumit Mamtani}

\begin{document}
\maketitle

\section{Abstract}
This project aims to explore the process of deploying Machine learning models on Kubernetes using an open-source tool called Kubeflow \cite{kubeflow} - an end-to-end ML Stack orchestration toolkit. We create end-to-end Machine Learning models on Kubeflow in the form of pipelines and analyze various points including the ease of setup, deployment models, performance, limitations and features of the tool. We hope that our project acts almost like a seminar/introductory report that can help vanilla cloud/Kubernetes users with zero knowledge on Kubeflow use Kubeflow to deploy ML models. From setup on different clouds to serving our trained model over the internet - we give details and metrics detailing the performance of Kubeflow. 

\section{Background and Motivation}
With the increase in use of Cloud Computing and the emergence of Distributed Systems due to the shear size of data and traffic over the internet, containers have become very popular due to their ease-of-use and scalability properties. A large number of companies have invested heavily in the management and deployments of such containerized applications. Kubernetes is one such open source system for automating deployment, scaling, and management of these containerized applications. 

According to a 2021 report in CNCF \cite{slashData}, there are more than 5.6 million users of Kubernetes. Mix that with the crazy increase in companies investing in Machine Learning, the intersection of the two is where Kubeflow fits in perfectly. 

\begin{figure}[h]
    \centering
    \caption{Rapidly increasing popularity of Kubernetes}
    \includegraphics[scale=0.4]{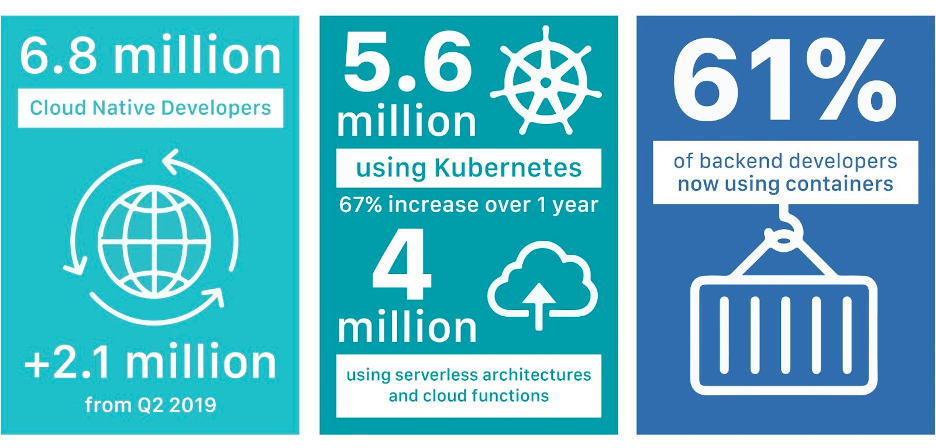}
\end{figure} 
Kubeflow lies in the intersection of Machine Learning, DevOps and Data Engineering - basically MLOps on the Cloud. 

\begin{wrapfigure}{r}{5.5cm}
\caption{What is MLOps}\label{wrap-fig:1}
\includegraphics[scale=0.3]{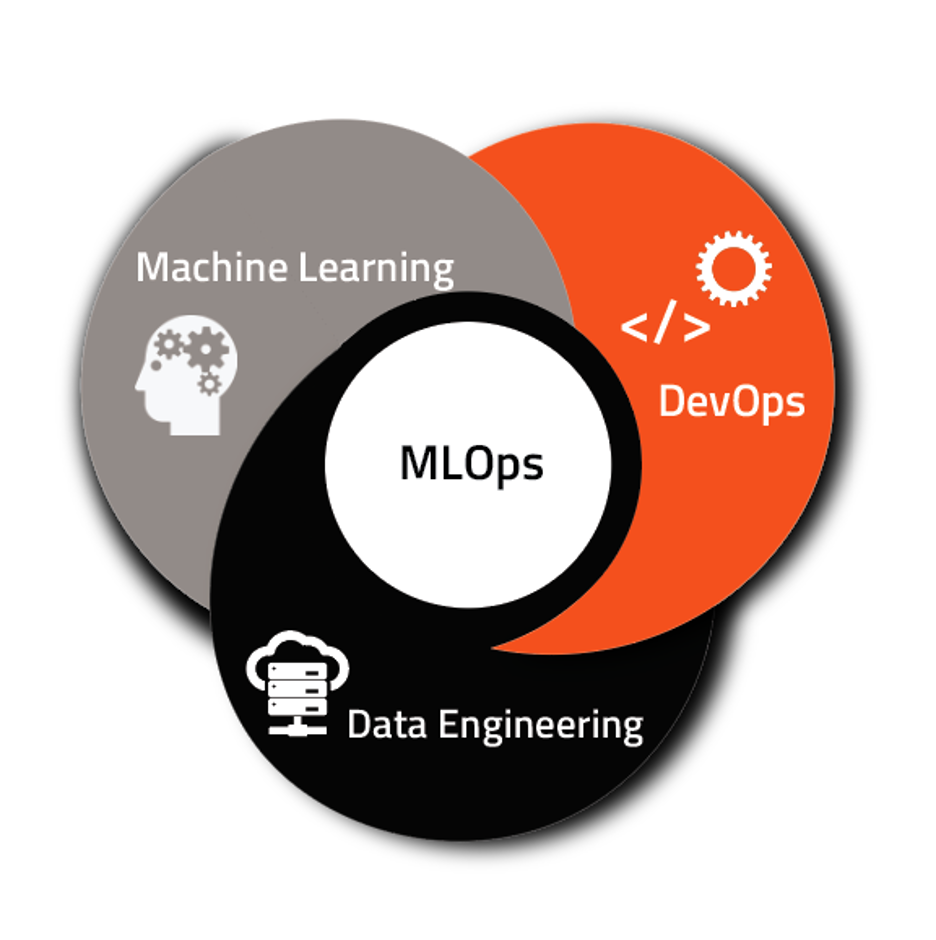}
\end{wrapfigure}

Kubeflow solves a number of pain points of MLOps - 
\\ (i) Inefficient tools and infrastructure
\\ (ii) Lack of iterative deployment 
\\ (iii) Importance of Automated CI/CD pipelines and 
\\ (iv) Handling growth of data and computing power.

Another important point to consider is the increasing complexity in deploying and maintaining large scale ML systems on the Cloud. While running ML models on developmental setups is straightforward and widely documented, productionizing them is a different ball game altogether. 

In the 2015 paper by Sculley et. al \cite{10.5555/2969442.2969519}, the authors talk about how the impression of considering the ease of building complex prediction systems using ML models is a free quick win is incorrect. The cost of being able to build such systems quickly is a large amount of \textbf{technical debt} in the form of maintanence costs of these real-world systems as well as several ML specific risk factors that need to be accounted for. 

Kubeflow helps us "pay off" this debt to some extent by standardizing and containerizing ML workflows \cite{payoff-arrikto}. 

\section{Related Work}
In terms of MLOps and generaline pipelining and orchestration platforms, there has been some related work. \\

Airflow is an open-source workflow management platform for data engineering pipelines. It was initially created by AirBnb \cite{airflow} to author, schedule and montior their workflows. Essentially, it is a generic task orchestration platform. While it is similar to
Kubeflow in many ways, Airflow was not built with Kubernetes in mind and is more useful for a generic use case. In fact, Airflow was initially not intended for ML pipelines at all and usually only performs orchestration and workflow management. \\

Argo \cite{argo} is an open sourcer container-native workflow engine for Kubernetes. It is again more of a general task orchestration problems that runs on Kubernetes natively. A part of Kubeflow is actually built on Argo. \\

MLFlow \cite{mlflow} is very similar to Kubeflow in the fact that it is a ML-focused workflow and pipeline management tool. MLFlow is supported by Databricks and is not constrained to Kubernetes and runs where the user chooses. It is more of an ML-focused tool and solves experiment tracking and model versioning. 

\section{MLFlow and Cloud Flavours}
\subsection{History}
TensorFlow \cite{TF} library was created by Google as an end-to-end platform for building and deploying  ML models. To orchestrate end-to-end TensorFlow Extended pipeline and run TensorFlow jobs on Kubernetes, Kubeflow was developed as an internal project (initially named TensorFlow Extended).

Kubeflow was open sourced at Kubecon in December 2017 and version 1.0 was announced on Febrauary 26, 2020 and since then, all the released versions are available on the its GitHub repository \cite{GH}. The idea was to offer ML model orchestration toolkit that can build on Kubernetes.

One big plus point of Kubeflow is that it is not locked into any Cloud Provider. Each provider provides similar services like Amazon SageMaker, IBM Watson, etc. but Kubeflow can abstract that to run on any cloud provider that supports Kubernetes. 

\subsection{Kubernetes Architecture \& Components}
Let's look at the main backbone and Kubeflow concepts we used - \\
\begin{figure}[h]
    \centering
    \caption{Kubeflow Components and Architecture \cite{kubeflow-arch}}
    \includegraphics[scale=1]{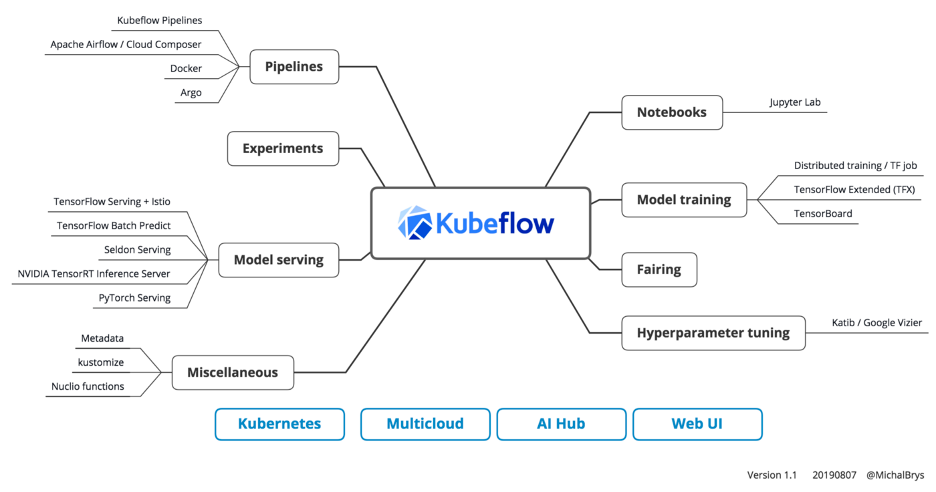}
\end{figure}
\\
\textbf{Pipelines} are the backbone of Kubeflow. They are used to build end-to-end ML workflows. The main goal of Kubeflow pipelines is to achieve End-to-end orchestration, enable easy experimentation and enabling easy re-use of components and pipelines to quickly create end-to-end solutions without having to rebuild each time \cite{kubeflow-pipelines}. \\
\\
\textbf{Notebooks} provide a way to run web-based development environments inside your Kubernetes cluster by running them inside Pods. Kubeflow provides native support for JupyterLab, RStudio, and Visual Studio Code. \\
\\
\textbf{Hyperparameter Tuning} in the form of \textbf{Katib} is a Kubernates-native project for automated machine learning (AutoML). Katib supports hyperparameter tuning, early stopping and neural architecture search (NAS) and can tune hyperparameters of many ML frameworks, such as TensorFlow, MXNet, PyTorch, XGBoost, and others. \cite{katib} \\
\\
\textbf{Model Serving} is to host machine-learning models (on the cloud or on premises) and to make their functions available via API so that applications can incorporate AI into their systems. For this, we use \textbf{KServe} (formerly KFServing) which is natively supported.

\subsection{Baseline Cloud Architectures} \label{base}
To establish a baselines to compare ease-of-use and performance aspects of Kubeflow, we run our training jobs on 2 types of platforms:
\begin{enumerate}
    \item \underline{\textbf{NYU Greene Cluster + Server}}: \\
    - This setup involved performing MNIST training on NYU Greene Cluster \cite{greene}, a HPC cluster to support research across wide range of implementations and disciplines. The cluster supports a range of job types and sizes requiring multiple CPU cores, GPU cards, TBs of memory or even a single core job. \\
    - It gives a flavour of trainig the Machine Learning code on bare metal and storing the model weights for further use. \\
    - For inference hosting, we used linserv machine \cite{linserv} that involves setting up an Apache Web Server. 
    - As this is the most basic setup we had, this is setup exactly lacks what Kubeflow is good for - automated pipelines. As we everything including environment setup/resource requesting needs to be done manually, this is a good baseline to compare the pros and cons of integrating Kubeflow. Also there is no Docker or Kubenetes support, unlike our next baseline.

    \item \underline{\textbf{Basic Kubernetes (on IBM Cloud)}}: \\
    - This setup involved developing Container and Kubernetes artifacts to perform Deep Learning model’s training and inference hosted on IBM Kubernetes cluster. To exceute the tasks, we need to create and deploy a Docker image to Docker Hub \cite{dockerhub} and then use the same image in our YAML files for creating jobs/containers. Hence, the MNIST model training and inference hosting was performed in the IBM Kubernetes cluster. \\
    - Unlike the previous setuop, here the environment and resource components are handled by Kubernetes and the setup does support generic containers, but we still needed to execute \textit{kubectl apply -f} command to deploy each YAML file to Kubernetes cluster. \\
    - What could be better is an end-to-end automated system that manages every stage from data preprocessing and training to inference, exactly what we will try to achieve using Kubeflow.

\end{enumerate}
With these baselines, we had the following two setups for testing Kubeflow:
\begin{enumerate}
    \item Running MNIST on Kubeflow in Google Cloud (GCP) - Using E2E and Code Approach
    \item Running MNIST on Kubeflow in IBM Cloud - Using E2E and Code Approach
\end{enumerate}
\subsection{Setup on Google Cloud}
While exploring the online documentation on the Kubeflow's official website, there were several possible paths to setup Kubeflow using Google Cloud \cite{KFGCP}, one of the approach involved creating Kubernetes Cluster on Google Cloud Platform (GCP), creating Notebook instance on GCP's AI Platform which could open JupyterLab instance and then creating a new instance of pipeline on same platform to finally link both so that code in notebook can create a new pipeline using the instance spawned earlier. \\
We were able to perform model training and testing on using this method, but due to out of date documentation and inadequate online documentation, could not proceed with inference of the model. Below are the screenshots of the procedure described above: \\ \\
\includegraphics[width=1\linewidth]{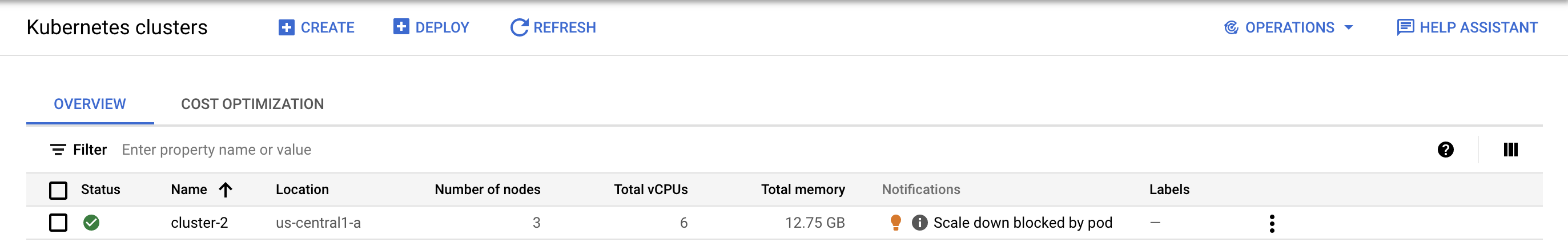}\hfill \\ \\
\includegraphics[width=1\linewidth]{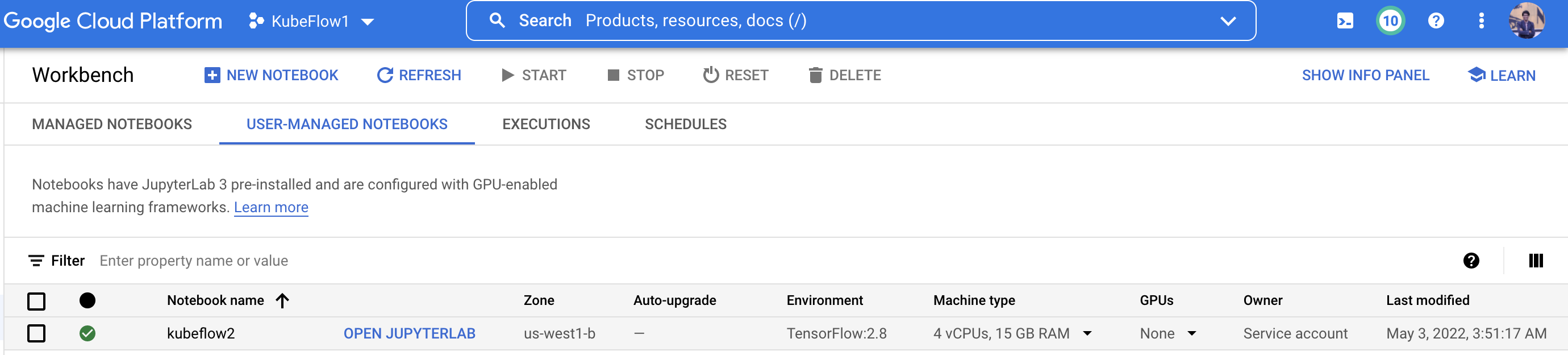}\hfill
\begin{figure}[H]
    \centering
    
    \includegraphics[width=1\linewidth]{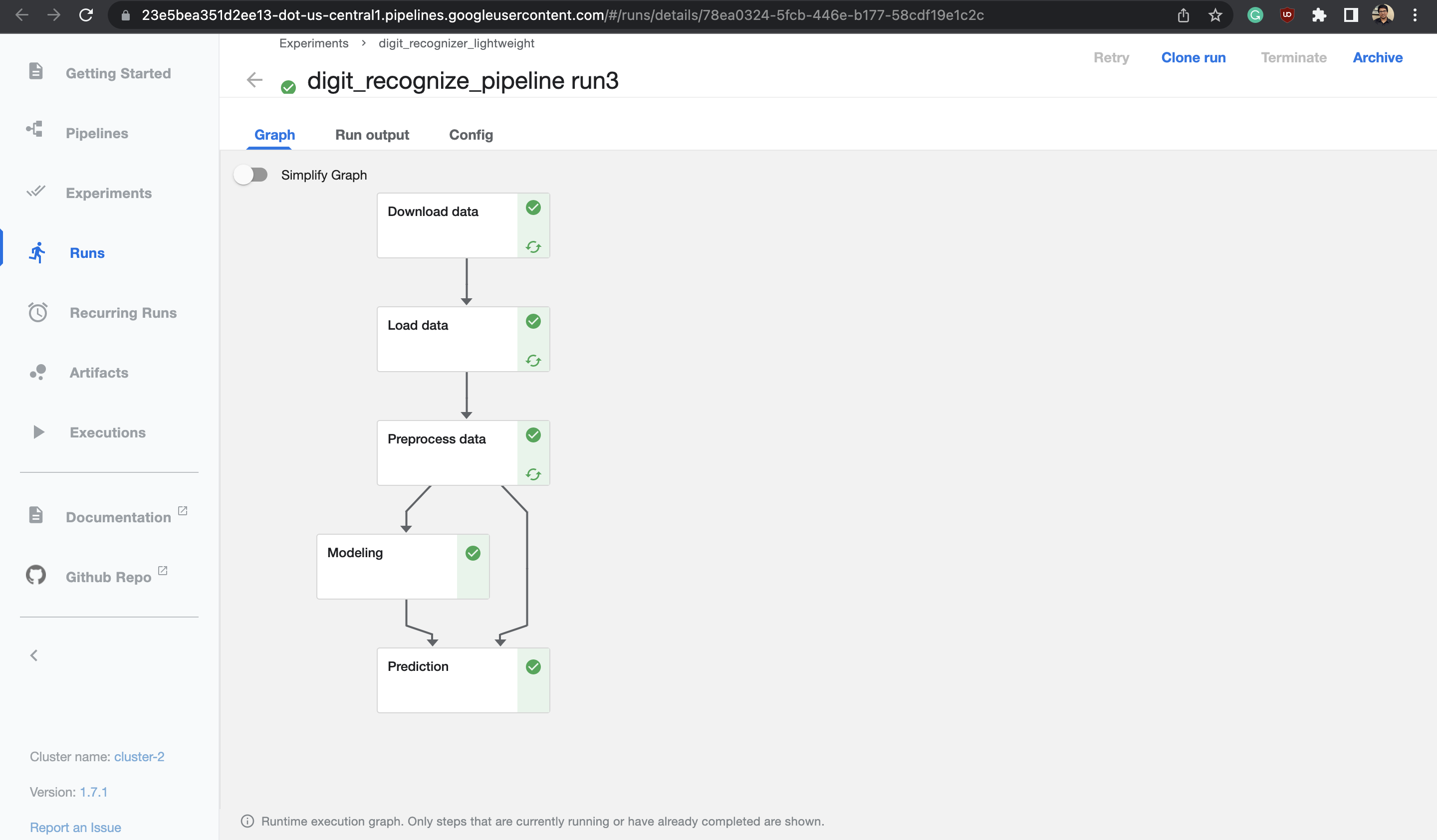}

    \caption{Creating KubeFlow pipeline on GCP - Attempt 1}
    \label{fig:kfgcpfailed}
\end{figure}
As one can notice in the pipeline image, there is no notebook option on the panel, which makes it difficult to setup the workflow as we need to create a notebook instance separately and then connect them via kfp.Client() - an API Client for Kubeflow Pipelines \cite{kfpclient} by providing the pipeline URL. The file - \textit{digit-recognizer-kfp-pipeline.ipynb} contains the code to to be executed on notebook instance to create Kubeflow pipelines on GCP. After facing several errors during inference stage, we decided to switch the methods. \\ \\
Surprisingly, the most easiest and resourceful way to create Kubeflow pipelines on GCP was not mentioned in most of the documentation - MiniKF - a single node, full fledged KubeFlow deployment. \\
Basically, MiniKF is to Kubeflow what Minikube is to Kubernetes. It is a single Virtual Machine solution that has capability to install Kubernetes, Kubeflow, Kale, Katib, KFServe, etc, necessary to train and serve your model. And hence, there is no requirement to create a separate Kubernetes cluster, Notebook instance or even a Pipeline instance, as everything will be available in MiniKF itself. \\
This is the method we used for creating our end-to-end Kubeflow pipeline on GCP. The steps are as follows:
\begin{enumerate}
    \item Create a project on GCP and ensure billing is enabled and IAM roles include editor privileges. Generally, if you are using your own GCP account and create a project, you would already be 'Owner' and have editor privileges for the project. 
    \item Using the search bar launch MiniKF from Marketplace. Here you will have to define Virtual Machine resources in ‘Configure and Deploy’ option. The creators of MiniKF recommend the following settings - 8 vCPU, 30 GB Memory, 200 GB SSD Boot Disk, 500GB SSD Data Disk. \\
    Although SSD memory is preferred to have, both for Boot and Data Disk, we experienced some problem while trying to create MiniKF instance with given settings (\textit{quota 'ssd\_total\_gb' exceeded}) and hence, switched to Standard storage for Data Disk, although it is slower than SSD, it is cheaper and was good enough for our experimentation. 
    \item After the instance is created, a window with an SSH button will appear with message 'Get started with MiniKF'. Once this appears, you can ssh in and type \textit{'minikf'} command and the rest setup will be carried out automatically. At last in the same window, a URL will be visible, were you can access the MiniKF dashboard using the username and password given. And that is it what all had to be done.

    \begin{figure}[H]
    \centering
    \begin{subfigure}{.4\textwidth}
      \centering
      \includegraphics[scale=0.2]{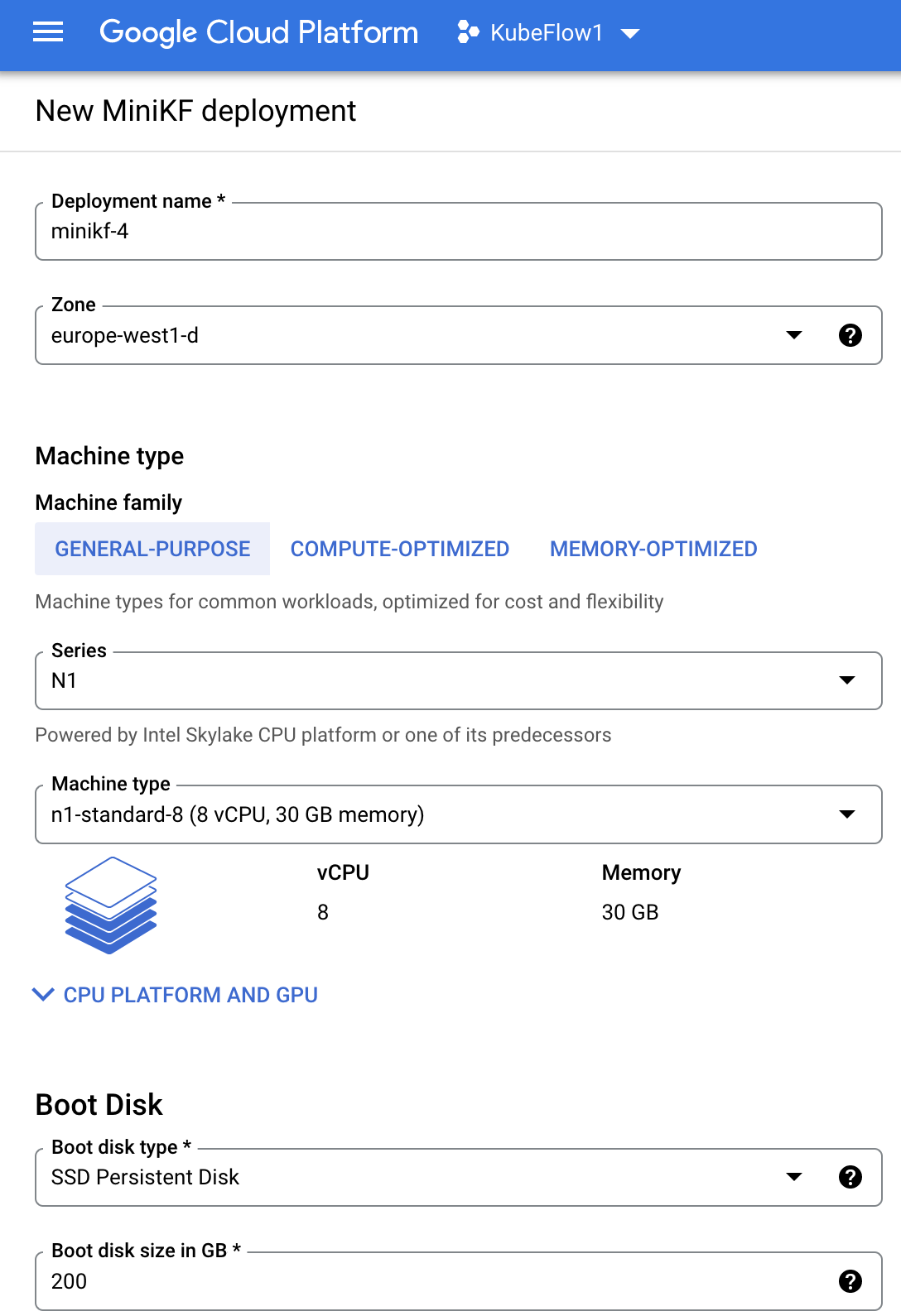}
      \caption{VM settings on MiniKF}
      \label{fig:sub1}
    \end{subfigure}%
    \begin{subfigure}{.6\textwidth}
      \centering
      \includegraphics[scale=0.26]{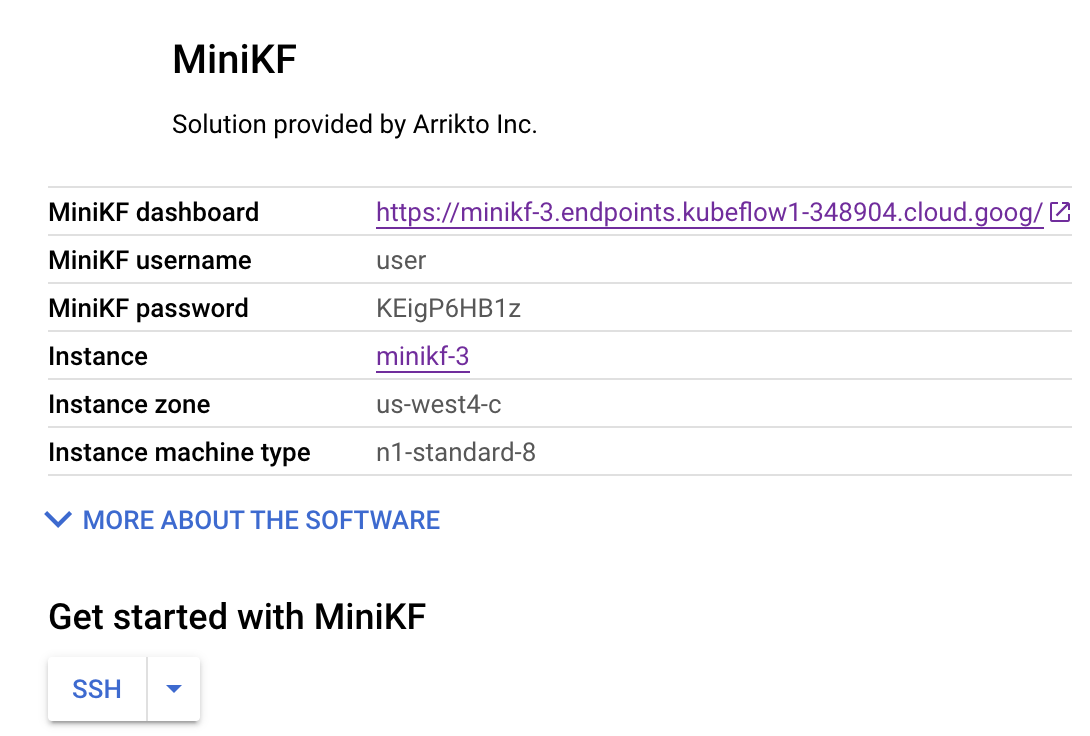}
      \caption{MiniKF instance ready}
      \label{fig:sub2}
    \end{subfigure}
    \caption{Creating MiniKF instance on GCP}
    \label{fig:minikfgcp}
    \end{figure}

\end{enumerate}
A summary of the steps followed can be found here - \url{https://youtu.be/porxGAcWnq8}. Visiting the URL for the MiniKF dashboard we see the UI which is very concise and easy to use. The left panel contains several options, one of which is 'Notebook' from where we can spawn a notebook instance directly into MiniKF instance and hence make use of the very functionality that was missing in the previous approaches. \\

\begin{figure}[H]
    \centering
    
    \includegraphics[scale=0.3]{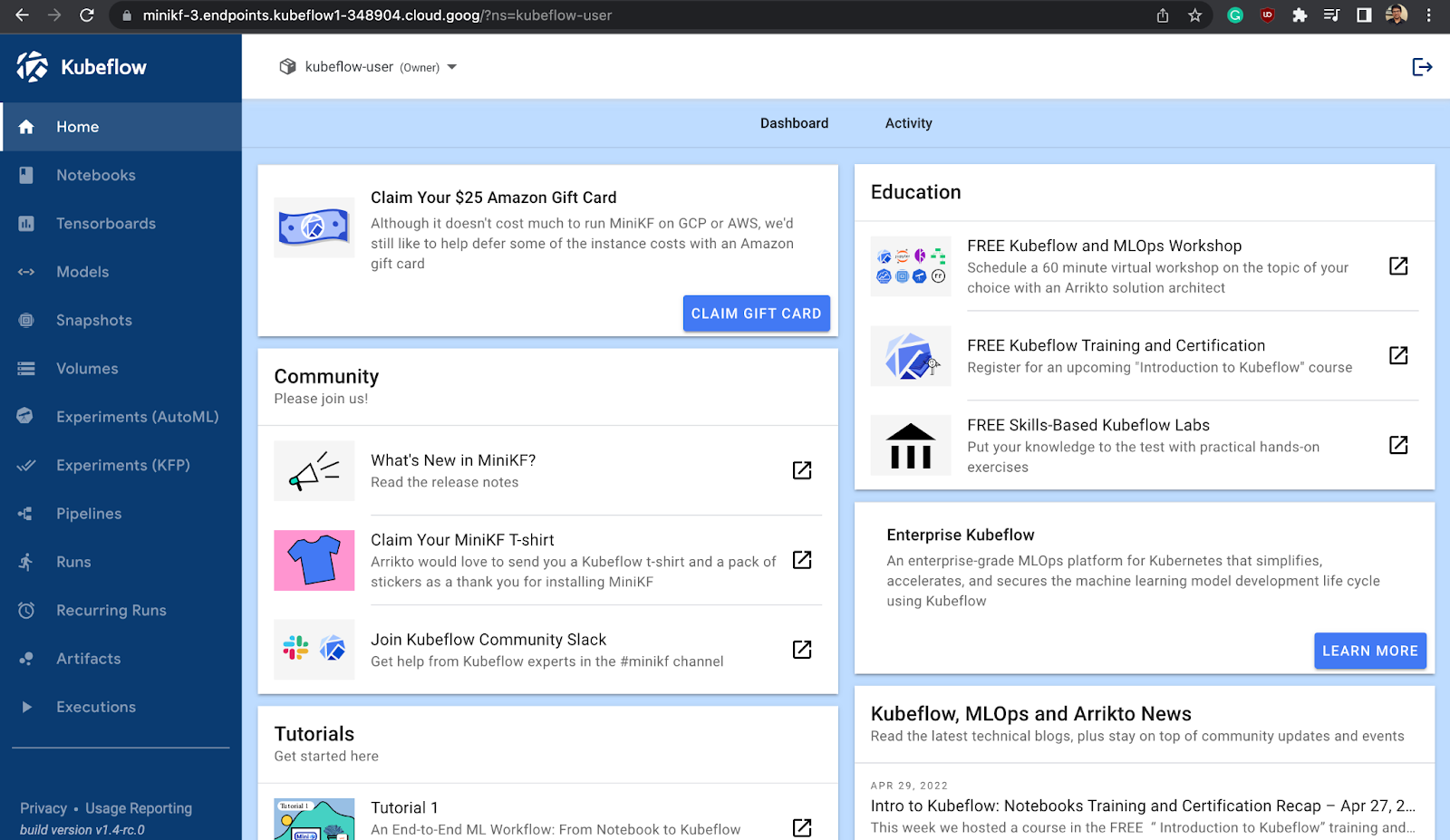}

    \caption{KubeFlow Dashboard on GCP}
    \label{fig:minikfdash}
\end{figure}

\begin{figure}[H]
    \centering
    
    \includegraphics[scale=0.25]{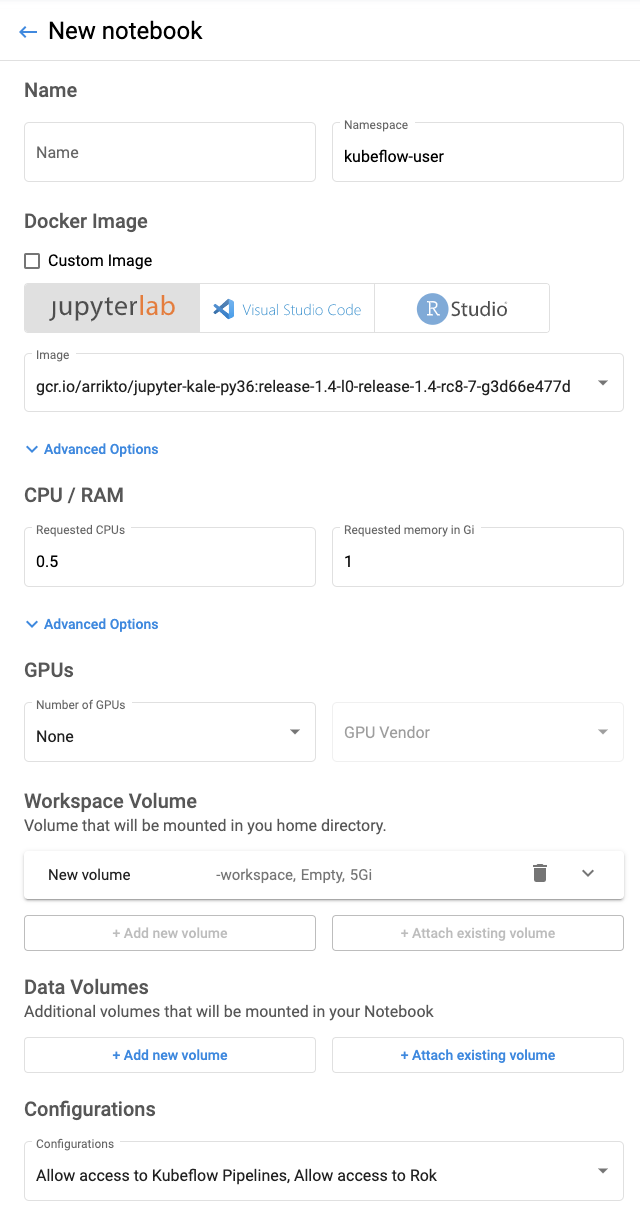}

    \caption{Notebook instance on Kubeflow Dashboard}
    \label{fig:notebookgcp}
\end{figure}

\subsection{Setup on IBM Cloud}

To set up Kubeflow on IBM Cloud, there were essentially 2 sources of documentation for the process - (i) on the Kubeflow Docs site \cite{kubeflow-docs-ibm} (ii) Provided by IBMs development team \cite{kubeflow-manifest} \\
\\
There are 2 main configurations of Kubernetes on which Kubeflow can be installed on IBm Cloud - \\
(i) Classic IBM Cloud Kubernetes cluster \\
(ii) vpc-gen2 IBM Cloud Kubernetes cluster \\
Both of these have different setup methods - especially for storage, authentication, network acces, etc. \\
\\
For this project, we have used a vpc-gen2 IBM Cloud Kubernetes cluster to deploy and setup Kubeflow. \\
\\
The steps to setup Kubeflow are as follows - 
\begin{enumerate}
    \item The first step is to create our Kubernetes cluster. \\
    While the step seems straightforward, there are a number of things that we need to keep in mind. \\
    - We need to make sure that the kubernetes cluster version is compatible with Kubeflow as Kubeflow on IBM Cloud is not compatible with the latest versions. \\
    - As we are using the VPC approach, we need to make sure that our VPC is setup and we create the cluster in the right location. 
    \item Next, we need to create the block storage and attach it to our subnet.  To create an effective Kubeflow platform, we set up/enable subnets, block storage and routing tables under the same region. 
    \item Now, we need to install \textbf{kustomize} \cite{kustomize} - which is a kubernetes native configuration management tool. Kustomize introduces a template-free way to customize application configuration that simplifies the use of off-the-shelf applications. This is on the same machine as the IBM CLI. \\
    - Note: Again the versions of kustomize need to be compatible. 
    \item Once we have the prerequisites, we can apply all the kubeflow configurations from the github link\cite{kubeflow-manifest}  above and our basic Kubeflow setup should be complete. The below three commands together will apply most of the configurations needed to install a basic flavour of Kubeflow. 
    \begin{lstlisting}[ showstringspaces=false, language=bash]
        $ git clone https://github.com/IBM/manifests.git 
        $ cd manifests
        $ while ! kustomize build example | kubectl apply -f -;
        do echo "Retrying to apply resources"; sleep 10; done
    \end{lstlisting}

\end{enumerate}
\begin{figure}[H]
    \centering
    \includegraphics[scale=0.35]{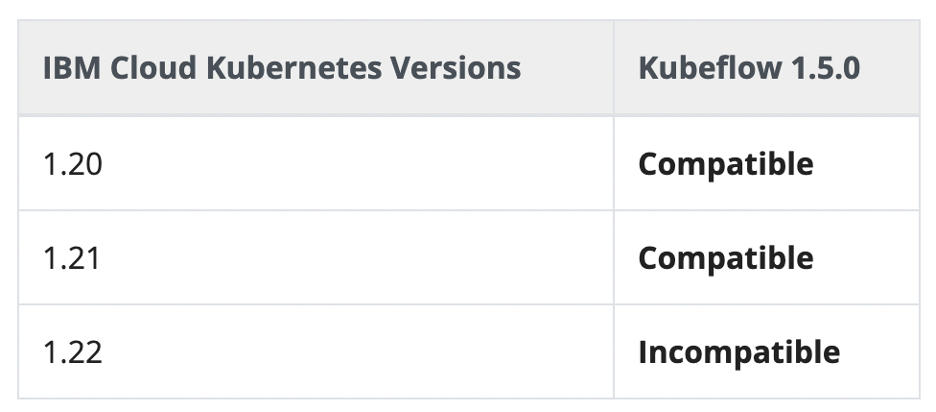}
    \caption{Compatibility of Kubeflow on IBM Cloud with Kubernetes}
    \label{fig:ibm_compat}
\end{figure} 
\begin{figure}[H]
    \centering
    \includegraphics[scale=0.35]{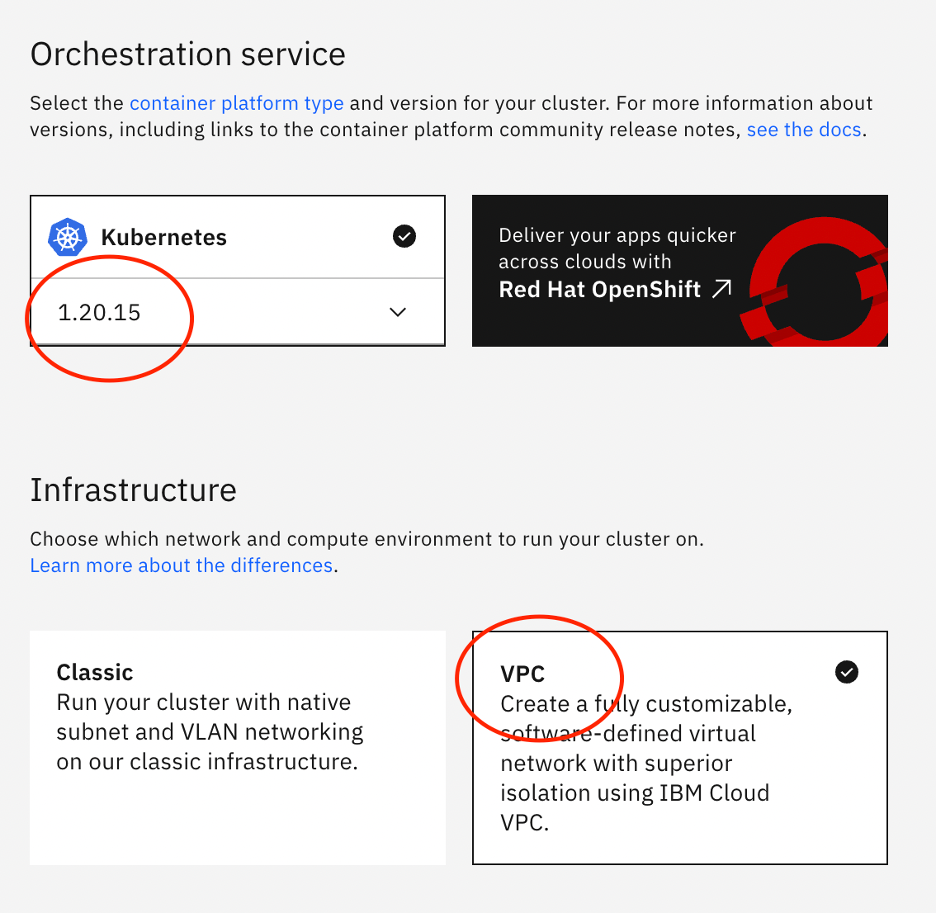}
    \caption{Choosing the Kubernetes version and Flavors correctly}
    \label{fig:ibm_choices}
\end{figure} 
Once we have our basic Kubeflow setup done, we can start playing with the features and access the dashboard. \\
But to have complete control over the platform, we need to do 2 further things - 
\begin{enumerate}
\item To view the Kubeflow dashboard, we just need to activate the ingress service using - \\
\textit{kubectl port-forward svc/istio-ingressgateway -n istio-system 8080:80} \\
The above command only works on the Kubernetes network. To expose it to the internet, we use - \\
\textit{kubectl patch svc istio-ingressgateway -n istio-system -p '{"spec":{"type":"LoadBalancer"}}' } \\

\item Although we now have the Kubeflow setup, the ingress gateways are only on HTTP. \\
To access Jupyter or any other Notebooks, we need to secure the ingress gateway endpoints with HTTPS. \\
For this, we follow the following steps - \\
    1. We create and setup a DNS for our Load Balanced endpoint\\
    2. We need to create a secret for the DNS and export it to the istio-ingress service\\
    3. Enable the kubeflow-gateway to use port 443 using the certificates\\
The steps are mentioned in \cite{ibm-civo} and \cite{ibm-auth}
\end{enumerate} 
We can not spawn notebooks, create pipelines and setup experiments on Kubeflow on our IBM Cloud Kubernetes Cluster. \\
\begin{figure}[H]
    \centering
    \includegraphics[scale=0.2]{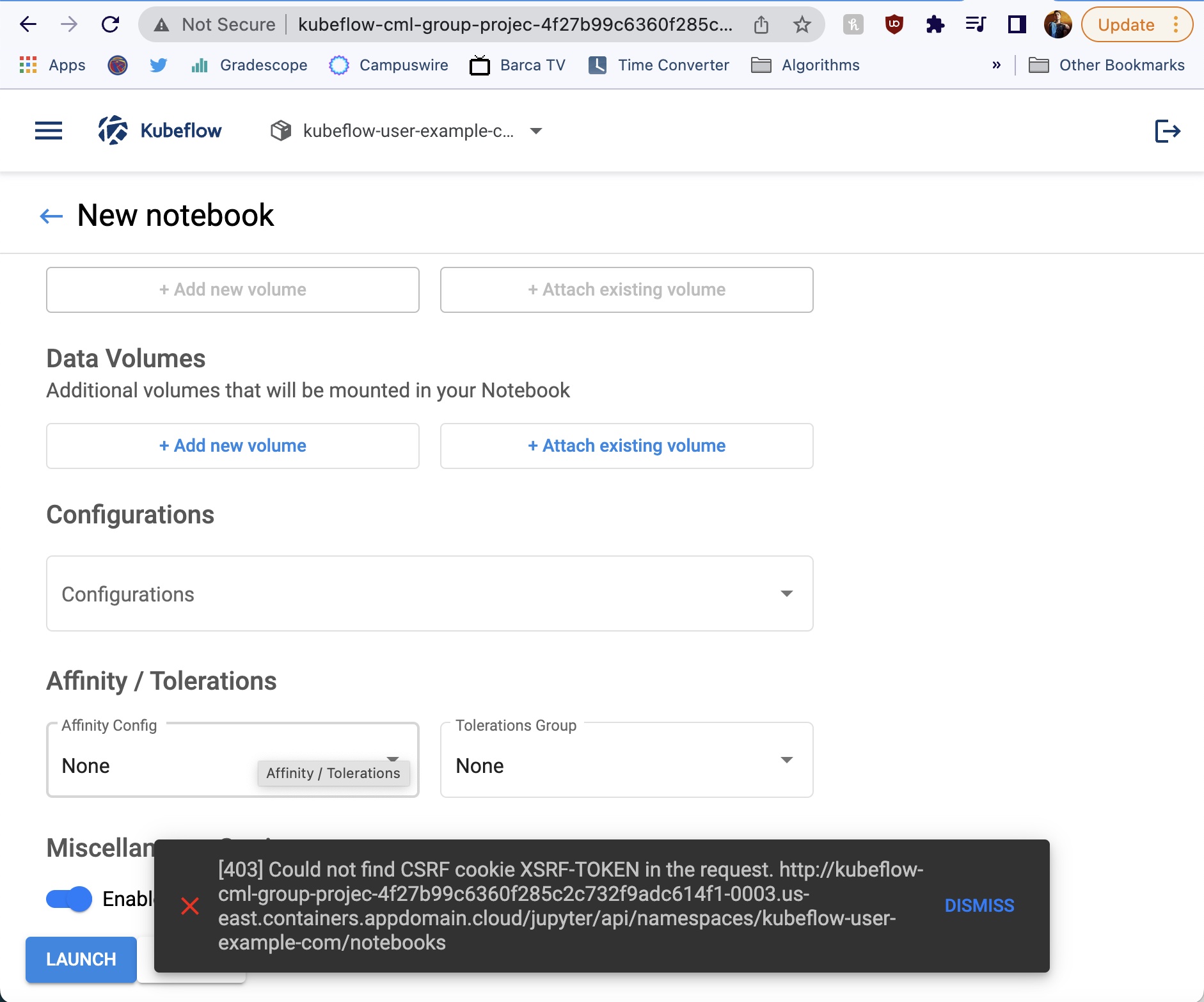}
    \caption{Notebook Creation failing on unsecured gateway}
    \label{fig:IBM_HTTP_Notebook_Fail}
\end{figure} 
\begin{figure}[H]
    \centering
    \includegraphics[scale=0.8]{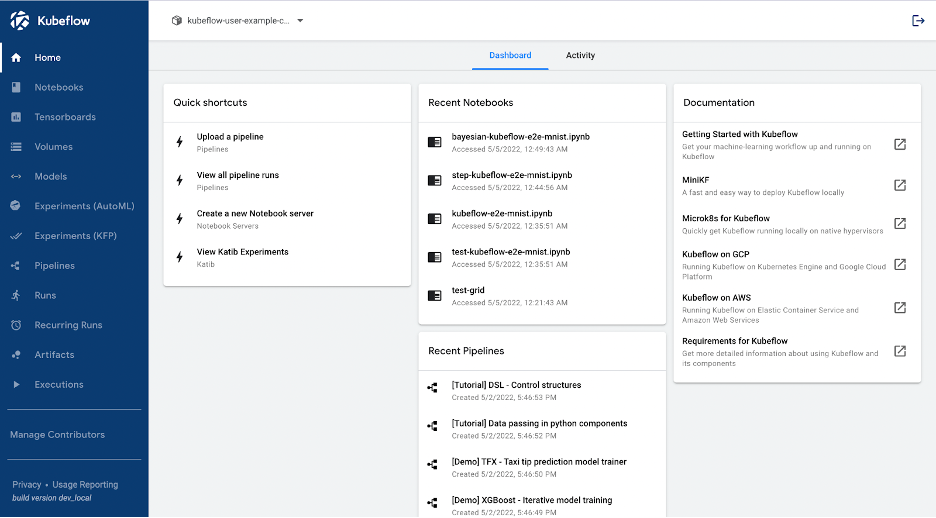}
    \caption{Kubeflow Dashboard on IBM Cloud}
    \label{fig:kubeflow_ibm_dash}
\end{figure} 

\section{Experimental Setup}
\subsection{Dataset}
For this project, we were more focused on the different features of Kubeflow and the end to end deployment of the ML Pipeline. \\
Therefore, we are using the MNIST dataset to experiment heavily with Kubeflow and different flavours of the same.
\subsection{Code Approach vs E2E Approach}
In terms of our code/pipelines we are creating and running on Kubeflow, we follow 2 main approaches - 
\begin{enumerate}
    \item Running a container image directly via a TFJob (End to End) \\
    In this, we run a Docker image containing a training code directly on Kubeflow and run a complete pipeline - right from data loading and preprocessing to model serving on KServe. This was on MNIST. \\
    The goal here is to showcase a complete end to end pipeling creation and execution on Kubeflow (including Hyperparameter tuning and AutoML) \\
    \begin{figure}[H]
    \centering
    \includegraphics[scale=0.5]{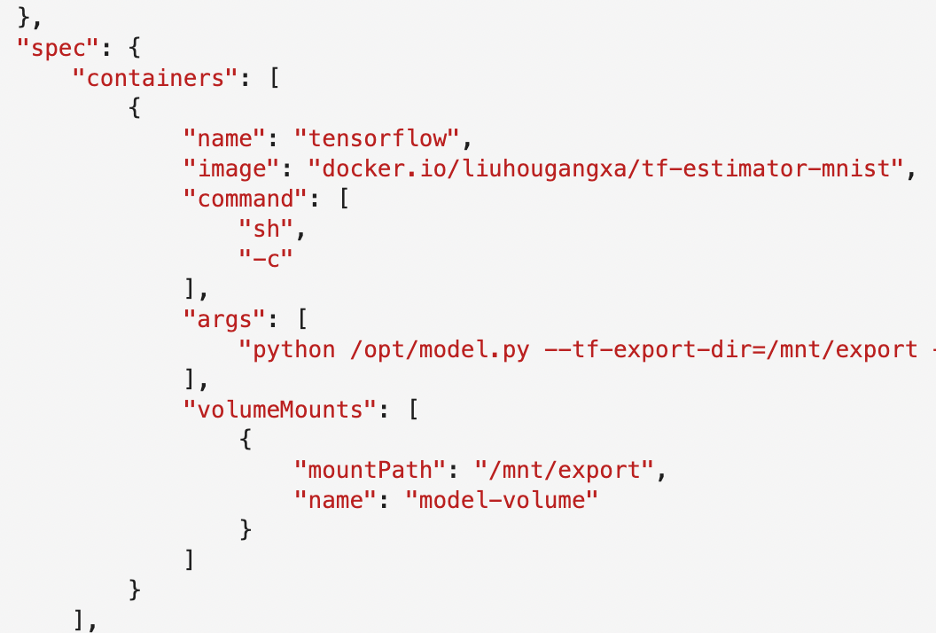}
    \caption{Running a direct image on Kubeflow}
    \label{fig:direct_image}
    \end{figure} 
    \item Creating a kubeflow pipeline by writing our own TF code over a base image. \\
    We also attempt to run a pipeline that contains a Neural network model defined by us by essentially converting our baremetal TF code to lightweight kubeflow component and launching it in a pipeline. This was also a Digit Recogoinzer on MNIST. \\
    \begin{figure}[H]
    \centering
    \includegraphics[scale=0.45]{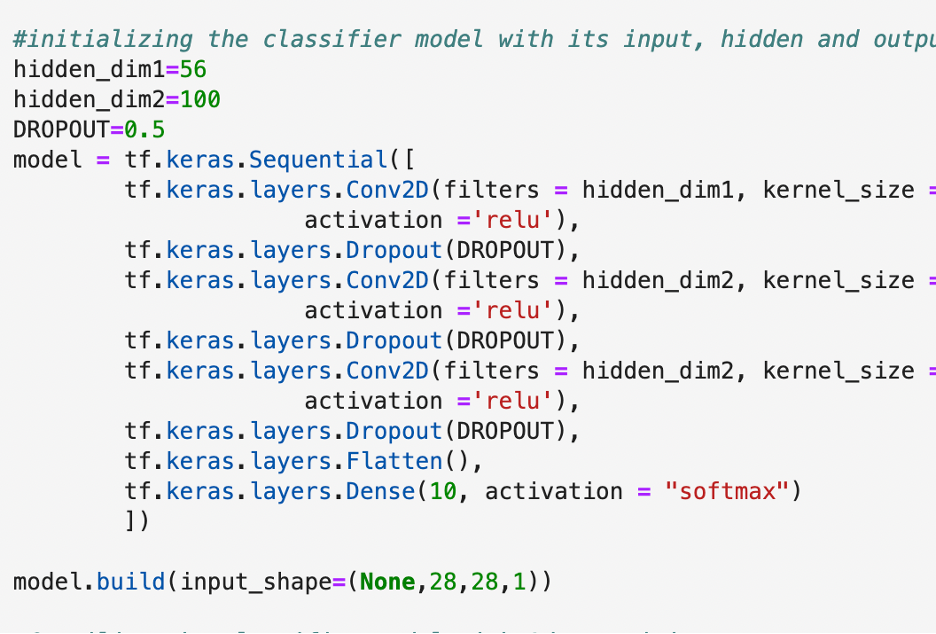}
    \caption{Custom Neural Network Model}
    \label{fig:custom_code_2}
    \end{figure} 
    \begin{figure}[H]
    \centering
    \includegraphics[scale=0.6]{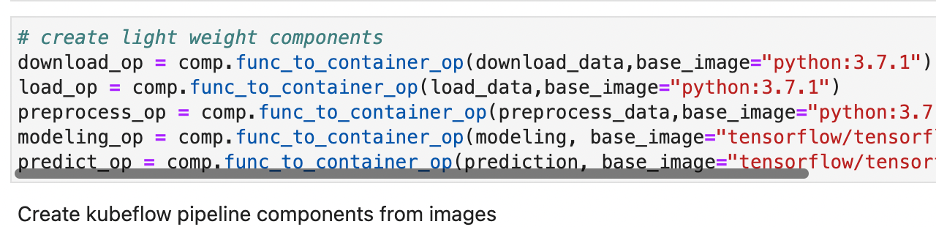}
    \caption{Using lightweight components to create a pipeline}
    \label{fig:custom_code}
    \end{figure} 
\end{enumerate}
Our goal here was to showcase the fact that users can indeed run their existing Machine Learning model on Kubeflow easily and without much modification while also having the flexibility of using Kubernetes's image pulll feature to train models. \\

\subsection{Pipeline}
Let's discuss the modelling of the pipeline that was used in the experimentation. 
\begin{itemize}
    \item \underline{Katib's Hyperparameter tuning task}: To dive deep into components of Kubeflow, we decided to explore Katib \cite{katib} - a Kubernetes-native project that can be used for hyperparameter tuning, early stopping, etc. Here we inspected the Hyperparameter tuning task. Objective was to minimize the loss while training the MNIST model and the goal was to reach $0.001$. The Docker image used in the code - \textit{"docker.io/liuhougangxa/tf-estimator-mnist"} uses LeNet -  an image classification ML model, to tune the hyperparameters. We decided to use random search over the hyperparamters which will search for the hyper-tuned parameters over a range of learning rate [0.01-0.05] and batch-size [80-100]. This algorithm will randomly choose over the values without replacement and hence will report the combination responsible for lowest loss. 
    \item \underline{TFJob Training Task}: This is a custom Kubenetes resource \cite{tfjob} that can facilitate running TensorFlow training jobs on Kubernetes instance created by MiniKF. This step will use the best hyperparameters found in Katib's experiment to train same model over which hyperparameters were tuned. 
    \item \underline{KServe Inference}: This resource \cite{kserve} is a standard Model Inference platform and can serve ML models on frameworks like TF, PyTorch, etc. This creates a serving component URL that will be used in inference of the model. The biggest advantage to use this asset is its autoscaling and intelligent routing capabilities for load balancing. This step also involves utilizing volume (defined in next step) as PVC, a persitant data storage for our ML model.
\end{itemize} 
At last, after creating a volume to load and store data generated while training and serving our model, we run the Kubeflow Pipeline with end to end MNIST model with hyperparamter tuning, training and inference. The code from the file \textit{gcp\_kubeflow-e2e-mnist.ipynb} shows that there is no need of any parameter for \textit{kfp.Client()} (no pipeline URL) as we are running notebook within the MiniKF instance. The run, when complete, is able to generate a single YAML file \textit{'minikf\_generated\_gcp.yaml'} that will guide the pipeline creation and hence the user can just code naturally to generate pipelines compared to writing a tedious YAML file all by themselves. \\
After running the notebook entirely we see the following outputted pipeline in our runs tab:

\begin{figure}[H]
    \centering
    
    \includegraphics[scale=0.3]{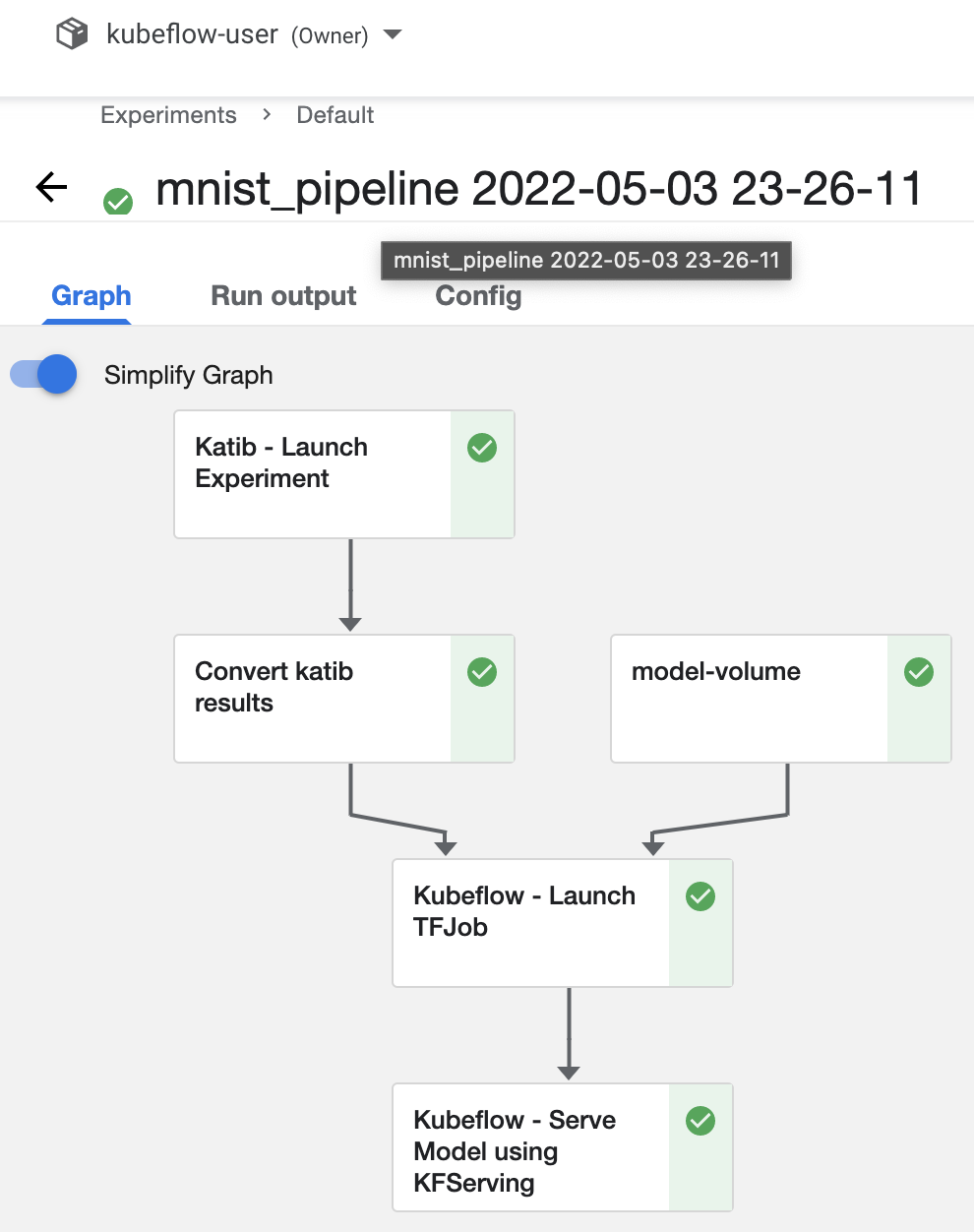}

    \caption{Successful pipeline completed after running the notebook}
    \label{fig:notebookgcp}
\end{figure}
For tuning, we used AutoML in the form of Katib integration with Kubeflow. We used random, bayesian and step algorithms. Based on all the trials we did, we obtained the following results:
\begin{table}[H]
\centering
\begin{tabular}{|c|c|c|c|}
\hline
\textbf{Platform} & \textbf{\begin{tabular}[c]{@{}c@{}}Best Trail \\ Performance\\ Loss\end{tabular}} & \textbf{\begin{tabular}[c]{@{}c@{}}Tuned \\ Learning \\ Rate\end{tabular}} & \textbf{\begin{tabular}[c]{@{}c@{}}Tuned \\ Batch-Size\end{tabular}} \\ \hline
\textit{IBM}      & 0.1876                                                                            & 0.453                                                                      & 92                                                                   \\ \hline
\textit{GCP}      & 0.2047                                                                            & 0.4980                                                                     & 93                                                                   \\ \hline
\end{tabular}
\caption{Table with hyperparameters tuned via Katib's experiments}
\end{table}
Under the 'Experiments (AutoML)' tab, we can access the information about the Katib's hyperparameter tuning as well, and here are the results for the experiments we performed:
\begin{figure}[H]
    \centering
    \begin{subfigure}{.5\textwidth}
      \centering
      \includegraphics[scale=0.15]{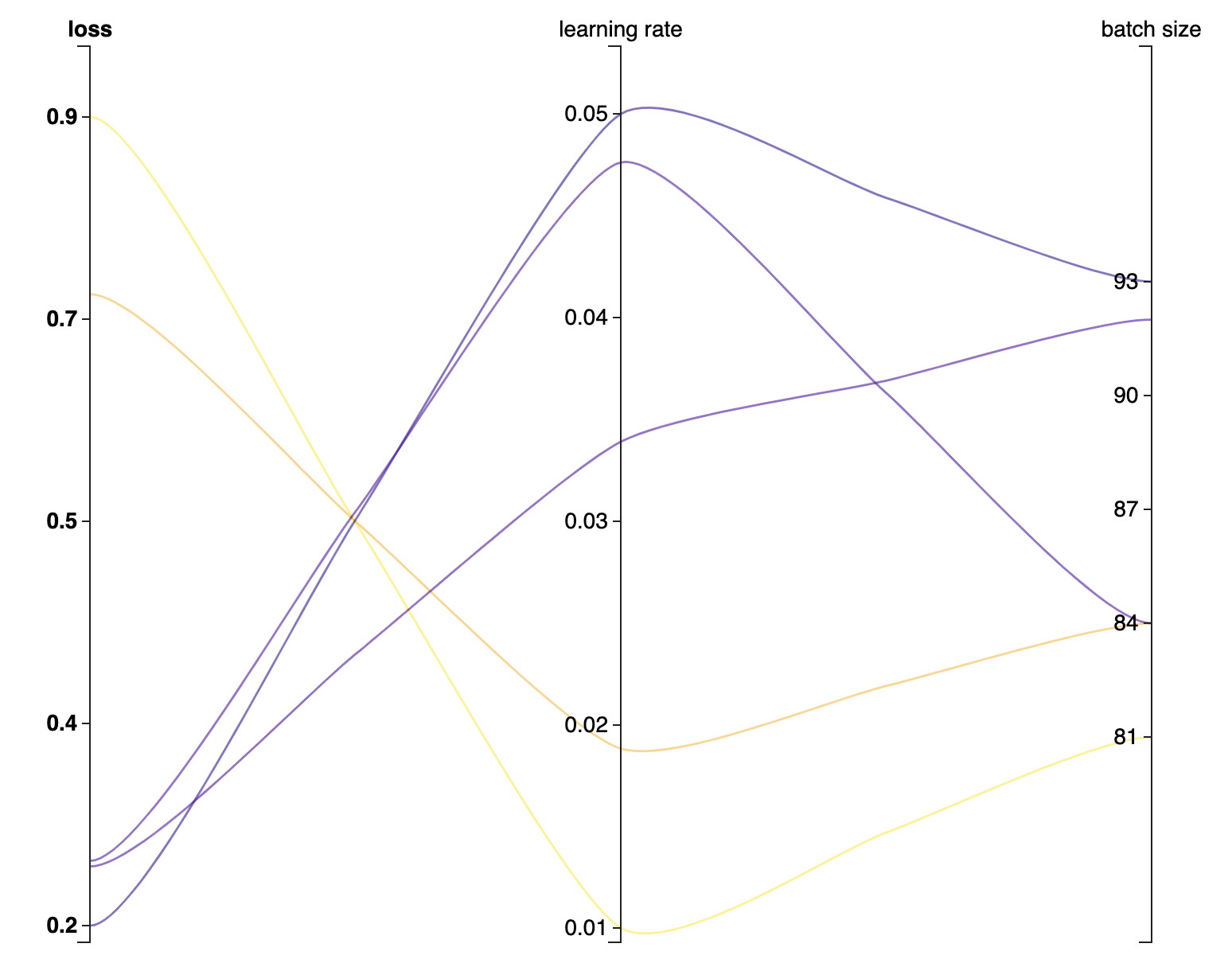}
      \caption{Katib on GCP}
      \label{fig:sub3}
    \end{subfigure}%
    \begin{subfigure}{.5\textwidth}
      \centering
      \includegraphics[scale=0.15]{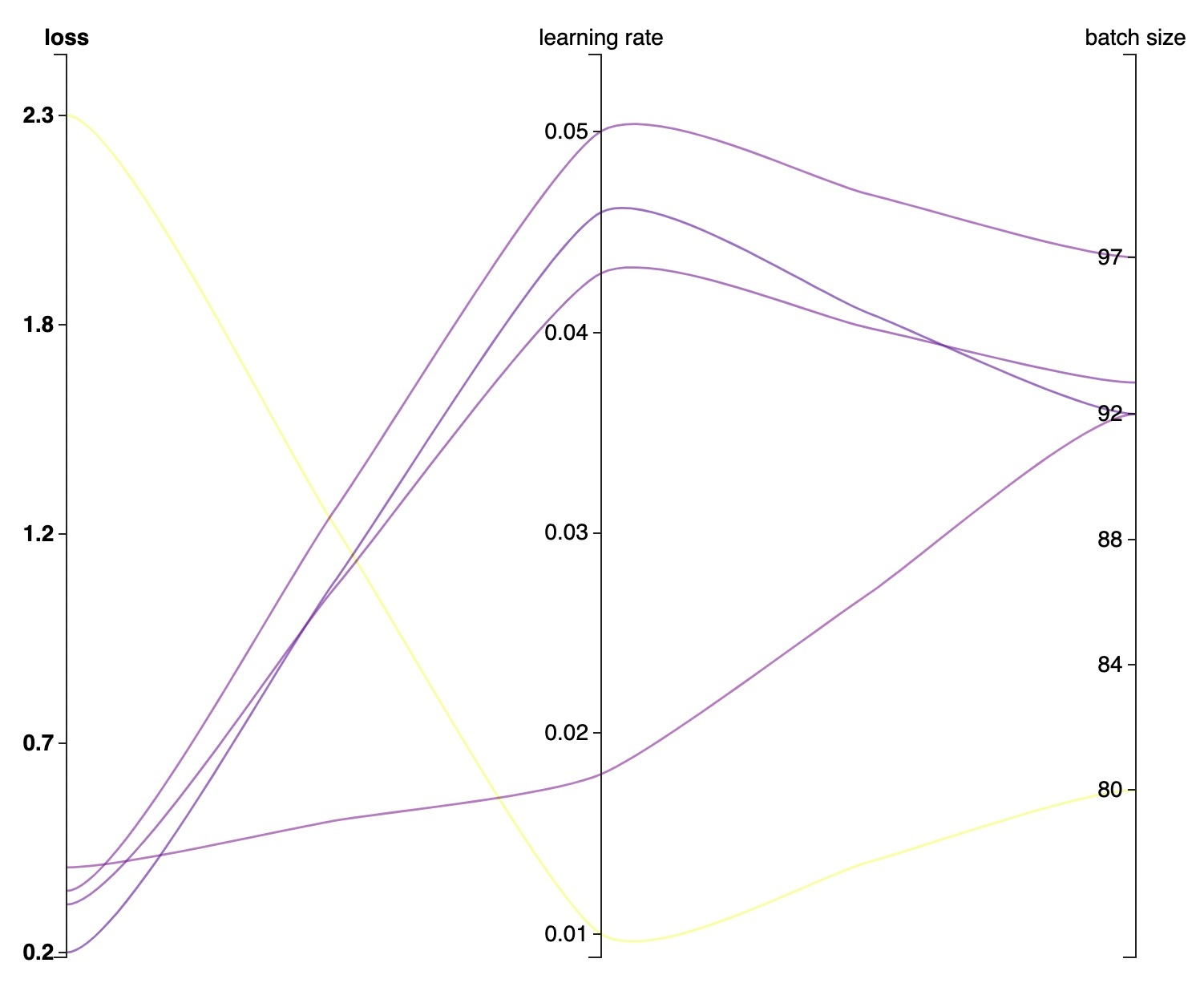}
      \caption{Katib on IBM Cloud}
      \label{fig:sub4}
    \end{subfigure}
    \caption{Katib's Hyperparmeter tuning on various Clouds}
    \label{fig:katib}
    \end{figure}

\subsection{Model Serving}
For model serving, we are using KServe to create our endpoints here. Kubeflow makes it extremely easy to create an enpoint and we can just pass our test values to the API endpoint using a CuRL request and get the result (as seen below).
\begin{figure}[H]
    \centering
    \includegraphics[scale=0.8]{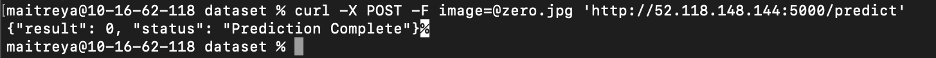}
    \caption{Serving through CuRL request}
    \label{fig:Serve_Curl}
\end{figure} 
This can also be done programatically using any request module - 
\begin{figure}[H]
    \centering
    \includegraphics[scale=0.8]{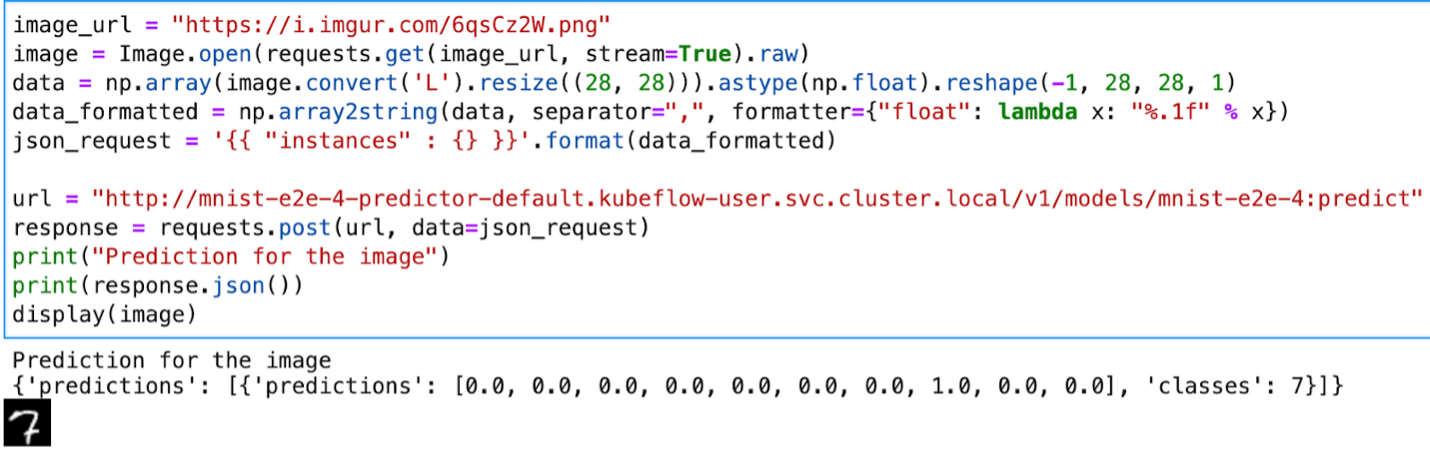}
    \caption{Serving through Python}
    \label{fig:Serve_Curl}
\end{figure} 
For the experiment running on linserv, we created a basic Flask-based app which loads a saved PyTorch model and makes a prediction and hosted it on the NYU Server. \\
\\
To showcase how an end-to-end ML based application might function, we also created a simple UI. The deployment for the UI Component is also very similar to any Kubernetes app deployment. \\
We create a simple static HTML page that lets users upload a photo and makes API requests to the inferencing API + displays the result.  \\
We then dockerize the html file and host is using \textbf{nginx} in a very lightweight form. \\
\begin{figure}[H]
    \centering
    \includegraphics[scale=0.3]{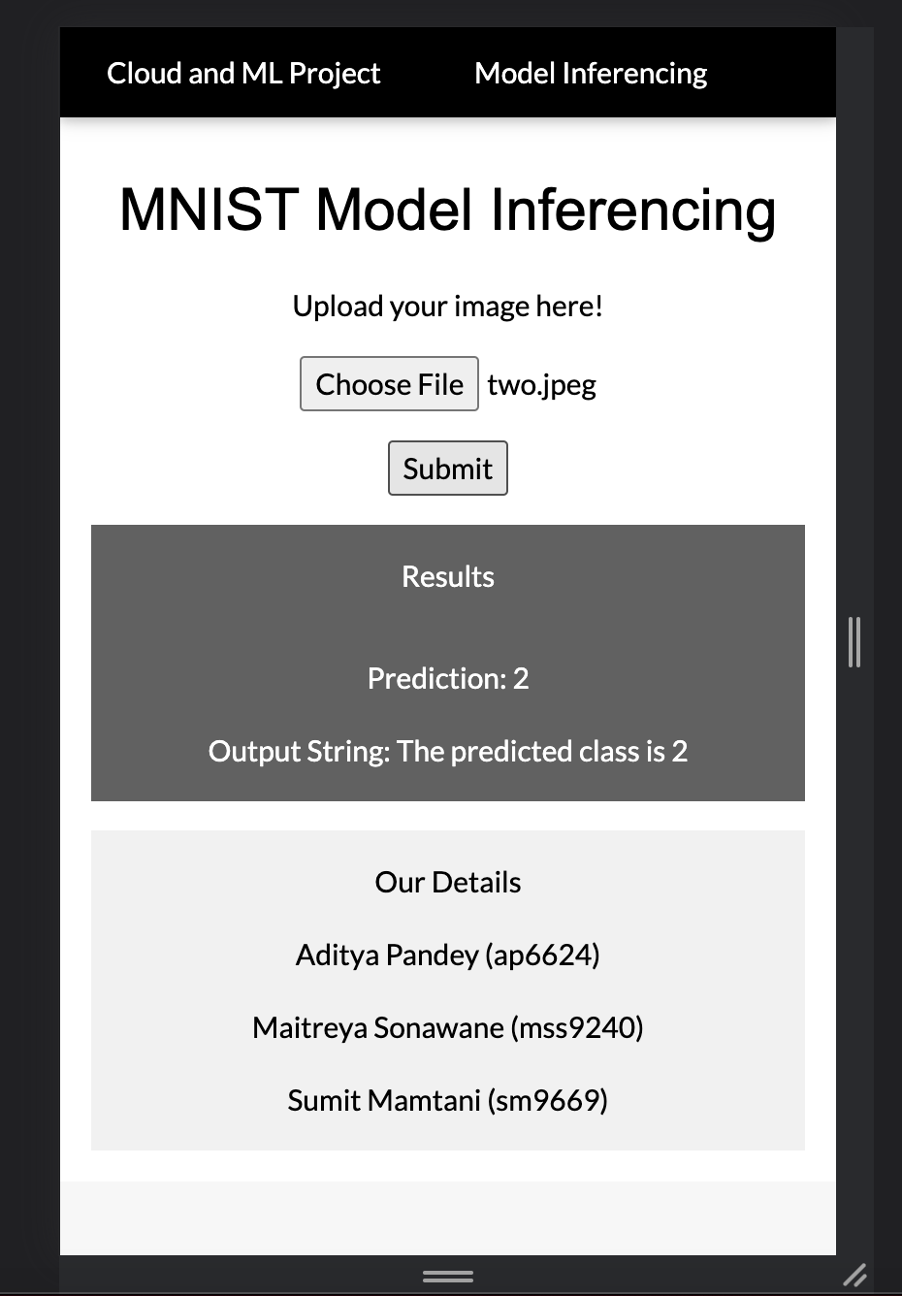}
    \caption{Simple Webpage making API requests}
    \label{fig:Serve_Curl}
\end{figure}
Once we have this docker image, we can now deploy our app to Kubernetes. We use a simple Pod to do this. \\
Now the last step is to expose this to the internet and we use a loadbalancer for this (similar to the previous part). Note that as there is only one pod, using a LoadBalancing Service is overkill but it provides a nice accessible endpoint for our webpage. \\

\subsection{Kubeflow Addons Used}
Let's discuss the additional tools that we integrated with a Kubeflow deployment.
\begin{itemize}
    \item \underline{Istio}: \\
    - Istio is an open source framework used by Kubeflow to enable end-to-end authentication and access control. It is a highly performant open-source implementation of service mesh used to describe the network of microservices and interactions between them. Kubeflow is a set of tools, frameworks, and services that work together to create machine learning workflows. These workflows are created by combining multiple services and components. Kubeflow provides the infrastructure that makes it possible to put these components together. \\
    - Kubeflow uses Istio for Securing service-to-service communication in a Kubeflow deployment with strong identity-based authentication and authorization. Its other requirements can include failure recovery, metrics, monitoring, and traces for traffic within the deployment including cluster ingress and egress.
    \item \underline{KServe}: \\
    - KServe formerly known as KFServing provides an inferencing service on Kubernetes and it also provides performant, high abstraction interfaces for common machine learning (ML) frameworks like TensorFlow, scikit-learn, XGBoost, and  PyTorch to solve production model serving use cases. It supports advanced deployments with canary rollout, experiments, ensembles and transformers. \\
    - KServe also supports a modern serverless inference workload with autoscaling, networking, health checking, and server configuration including advanced serving features like GPU autoscaling, scale to zero, and canary rollouts to ML deployments.

\end{itemize}

\section{Results \& Comparisons}
\subsection{Katib Hyperparameter Tuning}
\begin{figure}[H]
    \centering
    
    \includegraphics[scale=0.55]{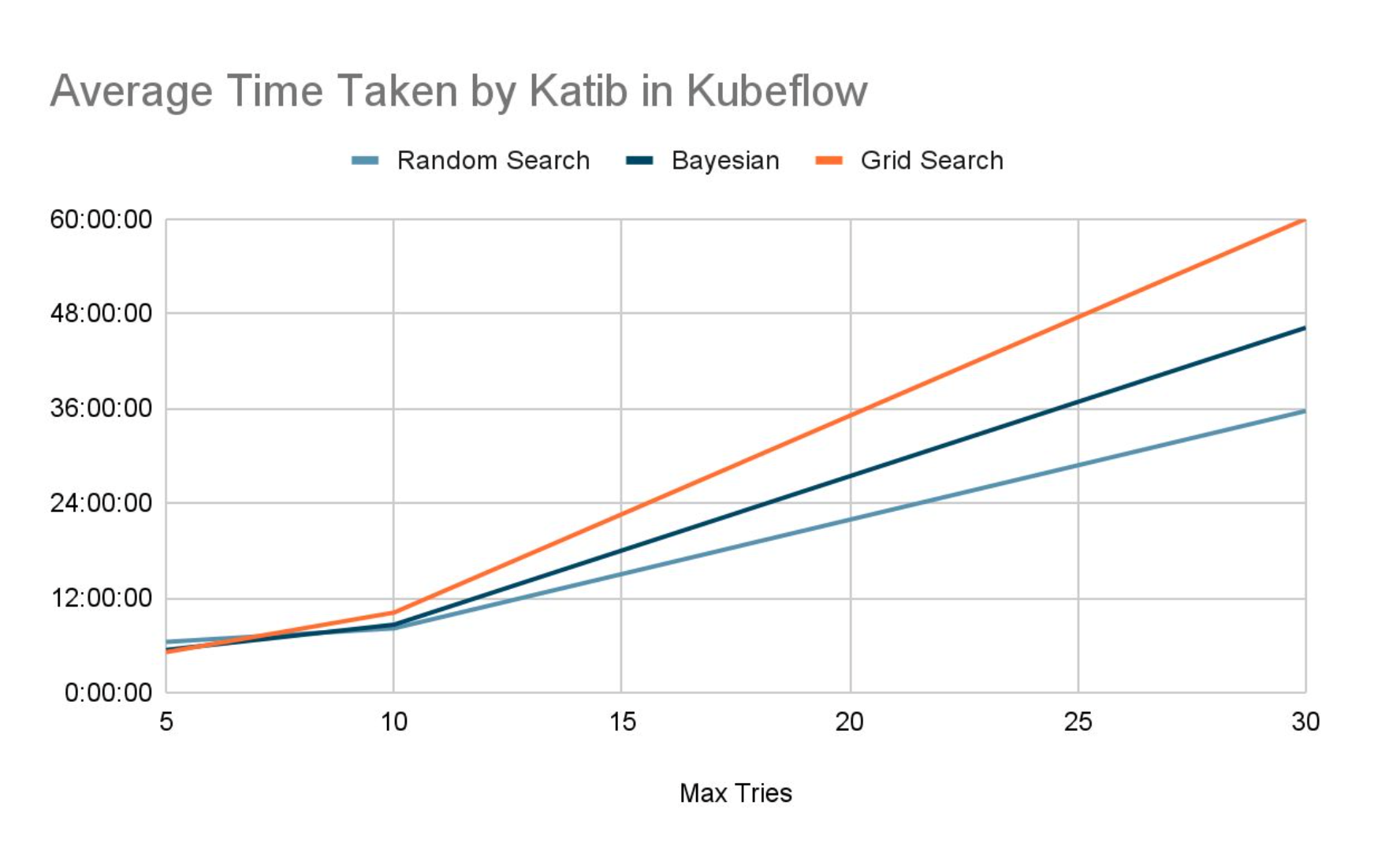}

    \caption{Average Time taken by Katib in Kubeflow for three Hyperparameter Tuning Algorithms  }
    \label{fig:search}
\end{figure}

We performed experiments on Katib, it is a Kubernetes-native project used for hyperparameter tuning, early stopping and neural architecture search (NAS). Experiments were done on Katib to find the optimum values of hyperparameters using three techniques - Grid Search, Random Search and 
Bayesian optimization search.
\begin{itemize}
    \item \underline{Grid Search:} Grid Search exhaustively searches this space in a sequential manner and trains a model for every possible combination of hyperparameter values. Grid search is not very often used in practice because the number of models to train grows exponentially as the number of hyperparameters is increased. This is very inefficient in time. We can also see from the Fig. \ref{fig:search} as we increase the number of runs the grid search is taking most time compared to Random and Bayesian optimization search.
    \item \underline{Random Search:} The key difference between random and grid search is that in a random search, not all the values are tested and values tested are selected at random. The advantage of randomized search is that we can extend our search limits without increasing the number of iterations (time-consuming) as we can see from the Fig. \ref{fig:search}. And the main point is that we can also use it to find narrow limits to continue a thorough search in a smaller region.
    \item \underline{Bayesian optimization Search:} Bayesian optimization is a sequential model-based optimization technique that uses the results from the previous iteration to decide the next hyperparameter values of the model. Bayesian optimization search is efficient because they select hyperparameters in an informed manner. When the model complexity is less Bayesian Optimization search takes less iteration to get to the global optima while Random and Grid search might take a high amount of iterations to get there. In our case, We have a somewhat complex model to train so that's why random search is taking less time than Bayesian and Grid Search.
\end{itemize}

\begin{table}[!h]
        \centering
        
\begin{tabular}{|c|c|c|c|}
\hline 
 Max Tries & Random Search & Bayesian Search & Grid Search \\
\hline 
 5 & 6:25:00 & 5:24:00 & 5:07:00 \\
\hline 
 10 & 8:08:00 & 8:36:00 & 10:08:00 \\
\hline 
 15 & 35:40:00 & 46:13:00 & 60:00:00 \\
 \hline
\end{tabular}
        \caption{\label{tab:table-name}This table shows the average time taken by Katib in Kubeflow for three hyperparameter tuning algorithms for different number of tries}
        \end{table}

\subsection{Comparison of Inference Time across all platforms}

We performed four different Experiments here.\\
Each experiment consisted of a model training instance and then an inference service hosted on the cloud platform. 
\begin{enumerate}
    \item We ran MNIST code on NYU Greene Cluster and deployed the model on Linserv for inference.
    \item Running a Basic MNIST Image on Kubernetes (on IBM Cloud) and then hosting a simple API with a LoadBalancing service
    \item Running MNIST on Kubeflow in IBM Cloud - Served using KServe
    \item Running MNIST on Kubeflow in Google Cloud (GCP) - Served using KServe
\end{enumerate}
To test the inference performance of each approach, we performed a sort of stress-test where we repeatedly sent predict requests consisting of one test image to the serving endpoint on the same network and noted the total time taken to respond to each request. Here are our results while comparing inference times on different platforms: 
\begin{table}[h]
\centering
\footnotesize
\begin{tabular}{|c|c|c|c|c|}
\hline
\textbf{\begin{tabular}[c]{@{}c@{}}No of\\ Req\end{tabular}} & \textbf{\begin{tabular}[c]{@{}c@{}}w/o KF \\ baremetal\end{tabular}} & \textbf{\begin{tabular}[c]{@{}c@{}}w/o KF \\ K8 cluster\end{tabular}} & \textbf{\begin{tabular}[c]{@{}c@{}}w KF \\ GCP\end{tabular}} & \textbf{\begin{tabular}[c]{@{}c@{}}w KF \\ IBM\end{tabular}} \\ \hline
1                                                            & 0.2415                                                               & 0.1955                                                                & 0.1194                                                       & 0.0603                                                       \\ \hline
4                                                            & 2.8616                                                               & 0.6382                                                                & 0.3362                                                       & 0.1974                                                       \\ \hline
8                                                            & 5.9860                                                               & 1.4011                                                                & 0.7692                                                       & 0.4774                                                       \\ \hline
16                                                           & 10.0755                                                              & 2.4590                                                                & 1.4794                                                       & 0.7617                                                       \\ \hline
32                                                           & 16.4755                                                              & 5.9566                                                                & 3.1525                                                       & 1.4677                                                       \\ \hline
64                                                           & 27.7447                                                              & 14.8050                                                               & 6.4079                                                       & 2.6950                                                       \\ \hline
100                                                          & 44.0205                                                              & 20.6780                                                               & 9.4831                                                       & 4.1400                                                       \\ \hline
128                                                          & 64.4582                                                              & 24.9734                                                               & 12.1261                                                      & 5.2376                                                       \\ \hline
256                                                          & 104.3657                                                             & 54.2752                                                               & 24.3457                                                      & 10.6503                                                      \\ \hline
512                                                          & 178.9776                                                             & 114.6778                                                              & 47.9780                                                      & 22.6391                                                      \\ \hline
\end{tabular}
\caption{Table of time required for inference on different platforms mentioned in \autoref{base}}
\end{table}
\begin{figure}[H]
    \centering
    \includegraphics[scale=0.42]{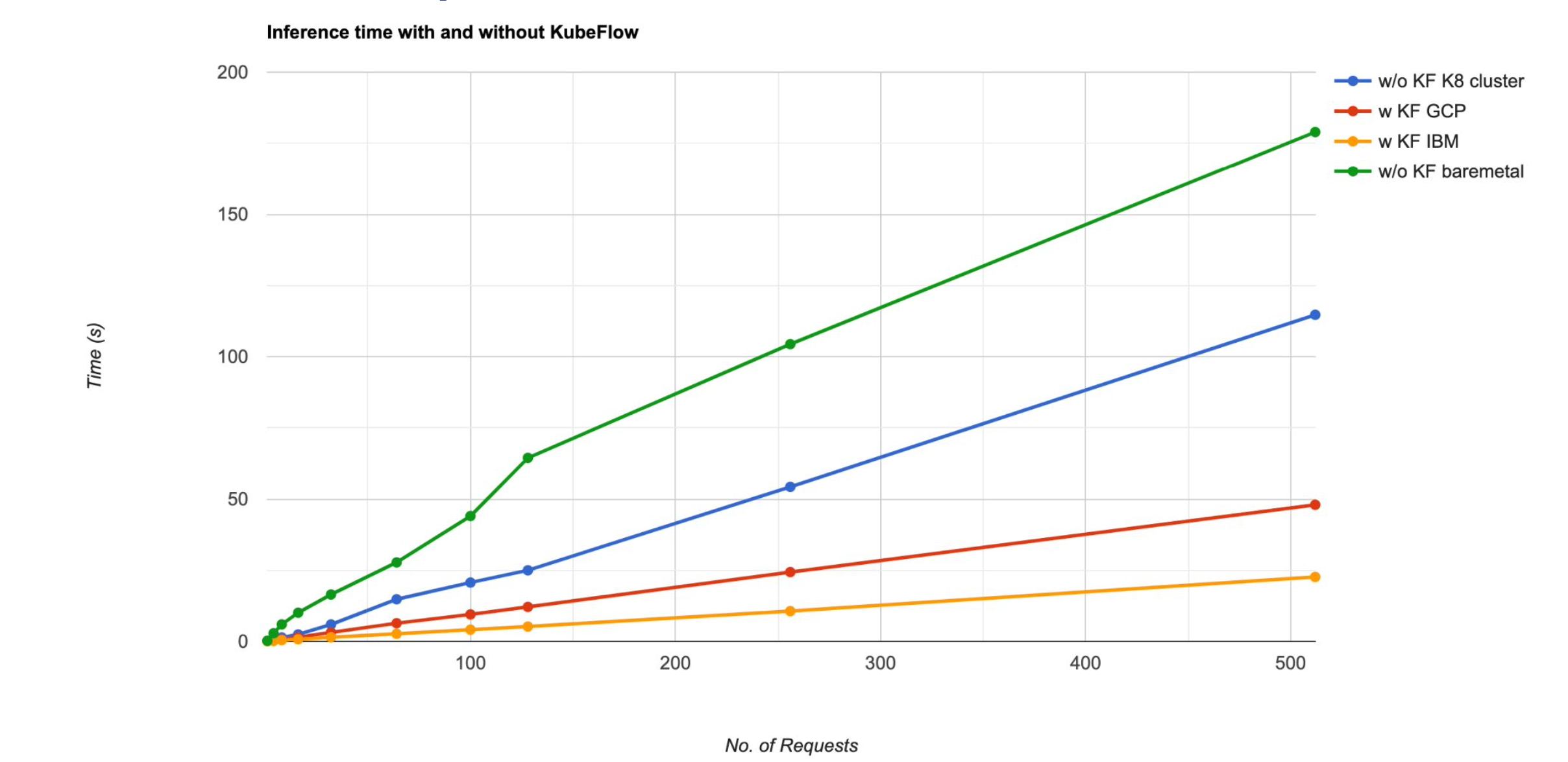}
    \caption{Inference Time Comparison with and without Kubeflow  }
    \label{fig:inference}
\end{figure}
We can see here that both the experiments run using the Kubeflow approach perform much better on the Stress test further showing us that Kubeflow is a great choice for model serving. 

\subsection{Comparing the performance of Kubeflow across clouds}
Our next experiment was to compare the performances of Kubeflow on the 2 different cloud providers - IBM Cloud and Google Cloud Platform. \\
We tested the running time of our 2 approaches - \\
(i) our custom model running a Digit recognizer \\
(ii) our E2E pipeline with Katib and Model Serving. \\
\begin{table}[h]
\centering
\begin{tabular}{|c|c|c|}
\hline
\textbf{}           & \textbf{Kubeflow on GCP} & \textbf{Kubeflow on IBM Cloud} \\ \hline
Total Pipeline Time & 745                      & 908                            \\ \hline
Model Running Time  & 429                      & 583                            \\ \hline
\end{tabular}
\caption{We compute the time required to run Kubeflow models on GCP vs IBM Cloud}
\end{table}

\begin{figure}[H]
    \centering
    \includegraphics[scale=0.9]{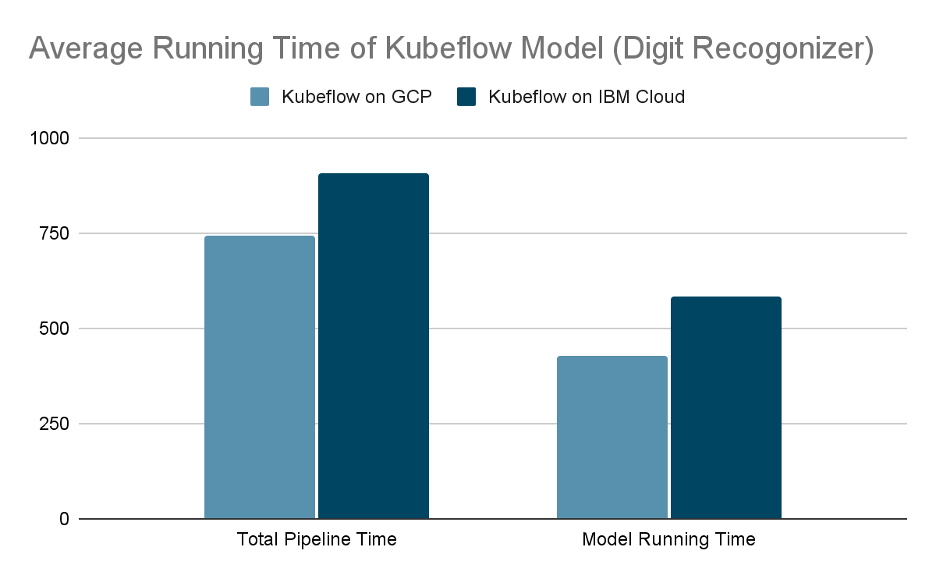}
    \caption{Running Time Comparison of our Custom Model }
    \label{fig:runtime_custom}
\end{figure}

\begin{table}[h]
\centering
\begin{tabular}{|c|c|c|}
\hline
\textbf{Time in sec} & \textbf{GCP} & \textbf{IBM} \\ \hline
Total Time           & 547          & 685          \\ \hline
Katib Experiment     & 338          & 385          \\ \hline
TFJob                & 126          & 108          \\ \hline
Model Serving        & 43           & 111          \\ \hline
\end{tabular}
\caption{Here we analyse the time required in every step for running Kubeflow E2E pipeline on GCP vs IBM Cloud}
\end{table}

\begin{figure}[H]
    \centering
    \includegraphics[scale=0.9]{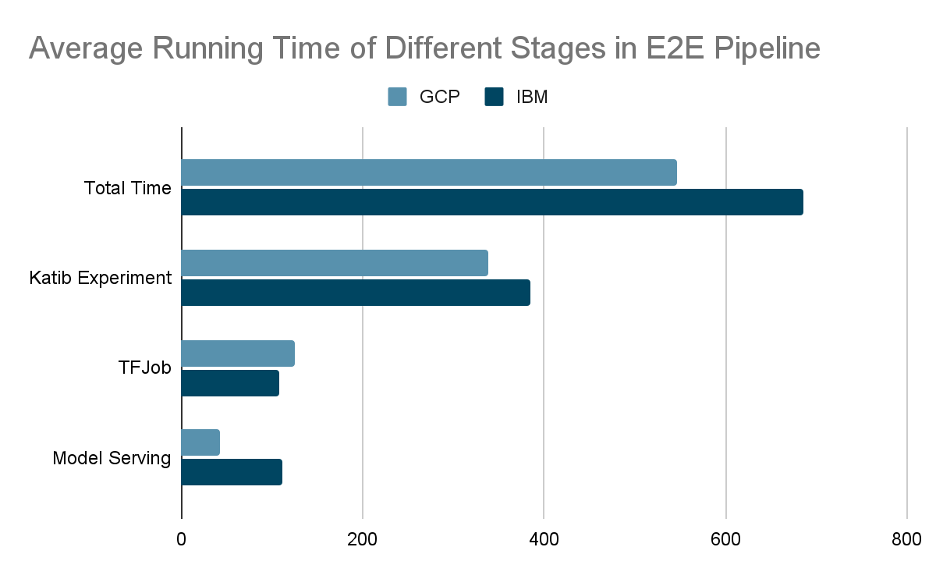}
    \caption{Running Time Comparison of different stages in the E2E Pipeline}
    \label{fig:runtime_e2e}
\end{figure}
We see that in both these approaches, Kubeflow on Google Cloud performs a bit faster than IBM Cloud on average. 

\section{Insights}
Based on the results above, we came up with the following insights - 
\begin{enumerate}
    
    \item
    Based on the sort of stress test on the inference endpoint for both clouds, we noticed that the MNIST inferentce model running on the Kubeflow on IBM cloud has the least inference time among all the models. \\
    One reason we think that IBM Cloud has a lower inference time is because all our K8s components on IBM Cloud are defined inside the same VPC in the same region. As we are making the network calls from the same network, a dedicated VPC would give us better network performance. \\
    \\
    The inference model API hosted on linserv is the slowest. This makes sense as the API is hosted on the public linserv server where we do not have any loadbalancing service and we are loading a new PyTorch model everytime a new request comes in. \\
    \\
    The API model hosted on Kubernetes makes use of loadbalancing features but still suffers from the same problem as above. 
    \\
    \item The Duration for running E2E pipeline is less for Kubeflow on GCP \\
    We feel that the total duration for running the pipeline on GCP is lower as the cluster is more powerful and the contention of resources is lower. \\
    While we tried to make the flavours of base compute the same for both clouds, the MiniKF setup on GCP is more idealistic and does the steup with minimal fuss. \\
    Another point here could be the fact that Kubeflow is a Google-based product due to which it's performance of GCP is slightly better. 
    \\
    \item The overall process of creating a cluster and using Kubeflow on it was easier on GCP for a few reasons - easier availability of documentation, automatic HTTPS endpoint securing, easy access to KF pipelines from Notebooks, etc.\\
    \\
    While IBM cloud has all these same features (as Kubeflow is platform agnostic), it is more challenging to enable/find resources or documentaion for the same.
    \\
    \item While Kubeflow has a lot of advantages, especially during Model training and inferencing, there are a few pitfalls that we feel prevent a more widespread adoption of the framework. \\
    \\ 
    The difficulty with the initial installation and authentication setup makes it a pain oint to start development with Kubeflow. \\
    As Kubeflow uses different versions of different components (such as Istio), upgrading individual components is a risky task. \\
    In addition, the presence of out of date documentation + Broken Links is a big challenge (as we talk about below).
\end{enumerate}

\section{Challenges}
Among the various challenges we faced during project, the most prominent ones were related to inadequate and outdated documentation of Kubeflow. The website \url{https://www.kubeflow.org} is mainly a documentation website that provides instructions on setting up Kubeflow and using it to achieve various tasks. Here are the major challenges we faced:
\begin{enumerate}
    \item \underline{\textbf{Deploying Kubeflow using UI}} \\
    There were so many resources that guide on setup and configuration of Kubeflow which start with the first step of Deploying Kubeflow using UI for example \cite{ner}. Here the problem is Deployment using UI is no longer supported for Kubeflow \url{https://www.kubeflow.org/docs/distributions/gke/deploy/deploy-ui}. The resources might be helpful enough to create Kubeflow environment but if the first step itself is broken, the resource is of no use. One should also note the inconsistency in the documentation as this website - \url{https://v0-6.kubeflow.org/docs/gke/deploy/deploy-ui} about Kubeflow v0.6 actually mentions that deployment can happen with URL, but takes us to a broken link when trying to open the documented URLs.
    \begin{figure}[H]
    \centering
    
    \includegraphics[scale=0.2]{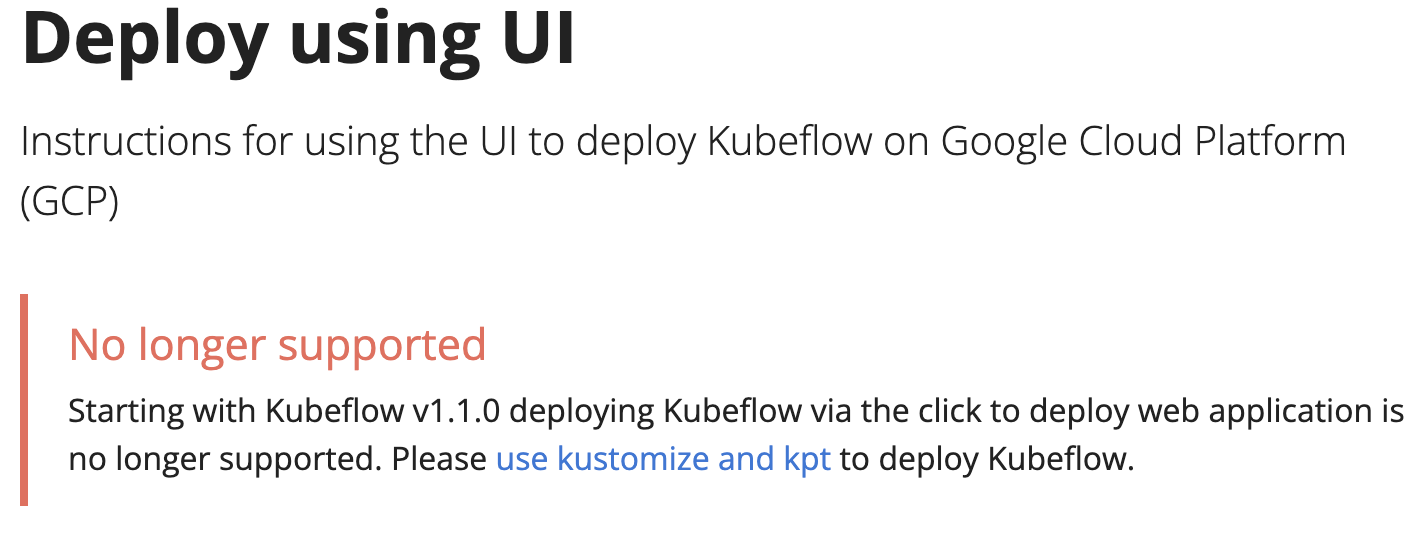}

    \caption{Deployment using UI no longer supported}
    \label{fig:deployui}
    \end{figure}
    \item \underline{\textbf{Component Specifications}} \\
    The Kubeflow Pipeline requires container component specification to describe data model. This will be ultimately serialized to a file in YAML format for sharing, similar to what can be found in the attached file \textit{'minikf\_generated\_gcp'}. As one can imagine, the step is one of the most important in the pipeline building phase, and yet the documentation if Out of Date - \url{https://www.kubeflow.org/docs/components/pipelines/reference/component-spec}.  
    \begin{figure}[H]
    \centering
    
    \includegraphics[scale=0.2]{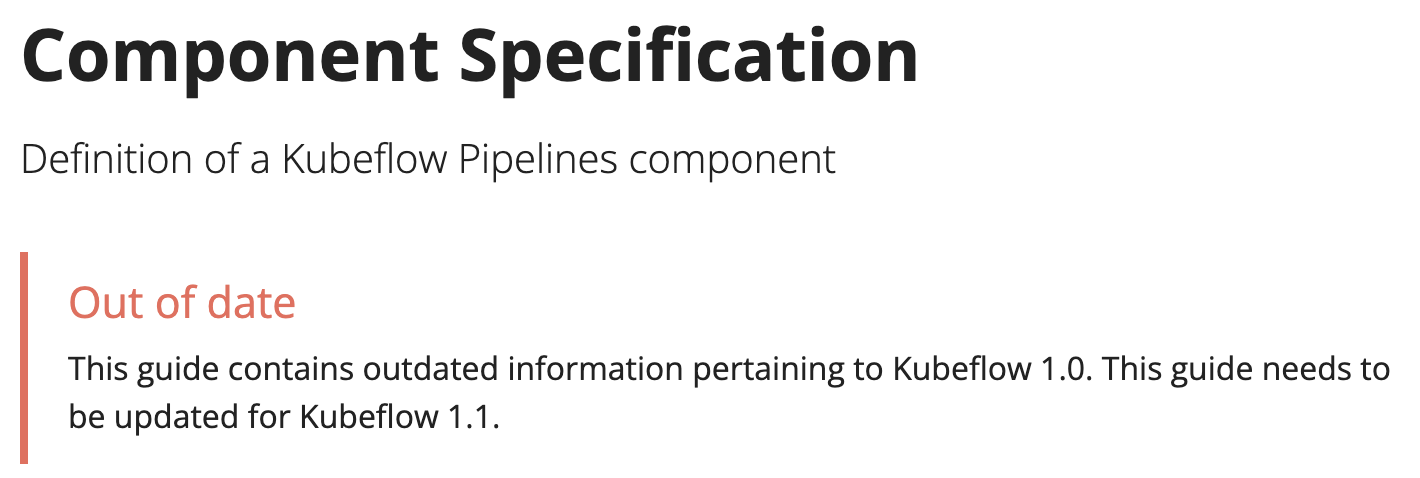}

    \caption{Component Specifications Outdated}
    \label{fig:compspec}
    \end{figure}
    
    \item \underline{\textbf{Official Documentation obsolete}} \\
    The model used in our experimentation for training, testing and inference was a ML model for MNIST dataset. The official repository of Kubeflow does provide an example of notebook to setup and run the same model. The problem - the documentation is outdated and the notebook is full of errors, possibly cause lots of component used are either removed or changed - \url{https://github.com/kubeflow/examples/tree/master/pytorch_mnist}. Remember, Kubeflow has been open sourced just in recent years. There have been several version rolled out in very short duration of time, for example, Kubeflow 1.0 in February 2020, Kubeflow 1.1 in June 2020, Kubeflow 1.2 in November 2020 and Kubeflow 1.3 in April 2021. Currently we are on version 1.5, and yet the documentation for the notebook is not updated since July of 2019.
    \begin{figure}[H]
    \centering
    \includegraphics[scale=0.3]{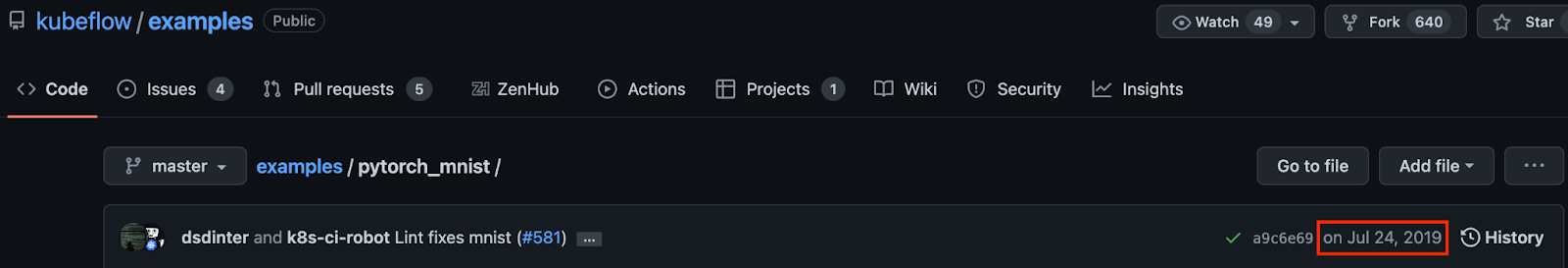}
    \caption{MNIST notebook obsolete}
    \label{fig:mnist2019}
    \end{figure}
    
    \item \underline{\textbf{Challenge of Setting up Kubeflow on IBM Cloud}}\\
    Setting up Kubeflow on IBM Cloud has a number of additional challenges than the above. \\
    The biggest one is the compatibility issues of installing Kubeflow on IKS. As the latest versions of Kubernetes and Kustomize are not supported, using existing clusters is a pain point as there is no clear way to downgrade the Kubernetes version. 
    \begin{figure}[H]
    \centering
    \includegraphics[scale=0.3]{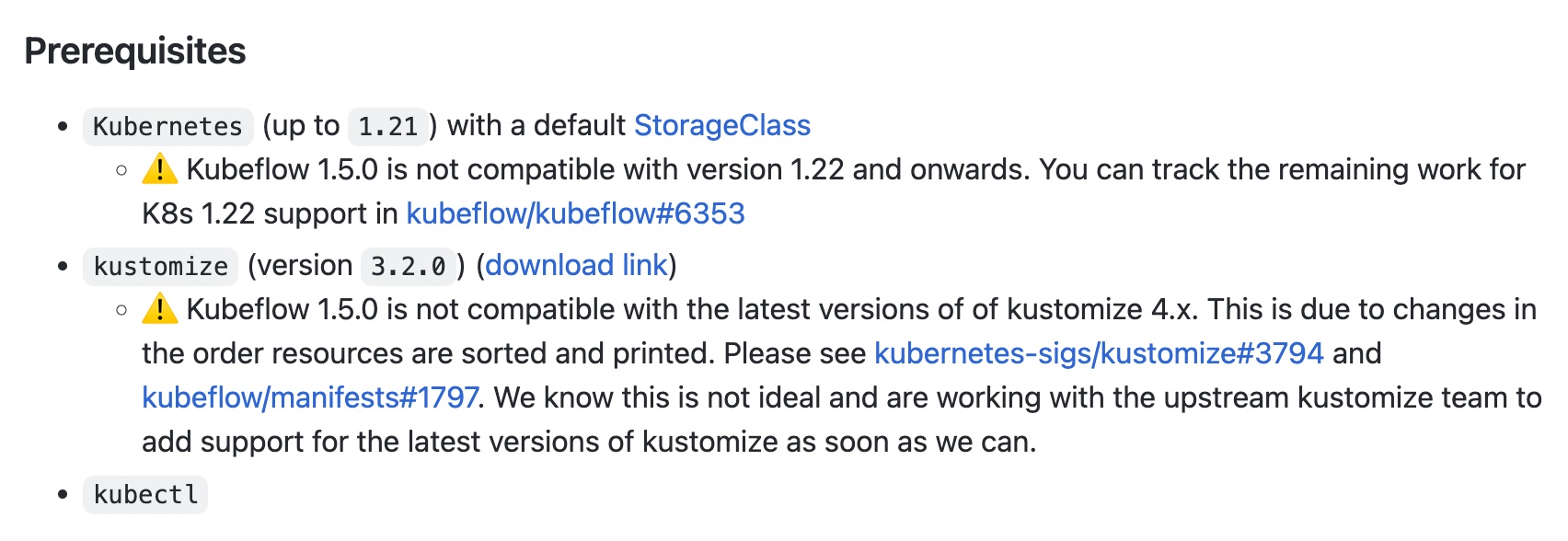}
    \caption{Lack of Support for latest versions}
    \label{fig:mnist2019}
    \end{figure}
    In addition, there are a number of broken links and the fact that there are not many public users of IBM Cloud definitely reduces the number of resources/support pages available to debug issues on Kubeflow on IBM Cloud.
    
\end{enumerate}
\section{Conclusion and Discussion} 
In this project, we presented a deep dive into integrating Kubeflow on both IBM Cloud and GCP, while comparing them with similar models deployed on K8s cluster and also models trained on NYU HPC without Kubeflow. \\
- We found that while duration of E2E run for Kubeflow on GCP was the least, IBM Cloud surpassed every other model to give fastest inference time. These results can be explained in the slight differences in setup and architectures of these 2 cloud providers. \\
- Kubeflow is a great tool and platform for users and developers to test and productionize end to end Machine Learning solutions. the ease of creating a pipeline - right from Data Ingestion/Preprocessing to Parameter tuning to Model serving makes it a valuable tool for any organizing wanting to deploy an ML solution. \\
- In addition, the presence of Kubernetes as a base makes use of all the advantages that Kubernetes provides in terms of Scalability and Orchestration. \\
But all this comes with a lot of effort needed to setup Kubeflow and replying on different services such as IStio, KServe, etc. that do have integration issues with the framework. 
- We must also note that While Kubeflow is great for running ML Jobs on Kubernetes; environments such as HPCs and Big Data Systems have their own flavours of MLOps \\
\\
We strongly believe that with better documentation and community support, Kubeflow has the right tools to be a successful framework. 


\section{Future Work}
We believe that there are many interesting avenues that can be explored with Kubeflow. \\
- For this project, we only explored CPU based execution of ML models due to limitations wth basic accounts on GCP and IBM Cloud. With more access (and credits), we could explore the integration of Kubeflow with GPU enabled Clusters and Notebooks. \\
- With GPU addition, we could also increase the depth of the neural network to see even more improvement over bare metal and increase number of trials in terms of Katib hyper-parameter tuning with different model architectures. \\
- One point that was mentioned was how Kubeflow was an extension of TensorFlow. But Kubeflow also supports PyTorch and other frameworks and this compatibility could be explored. \\
- We can also try an ML problem from a different domain (such as Speech/Text) and check if the advantages offered to us here also hold for those domains. \\
- Finally, we encourage anyone interested in this framework to try their hand at becoming an open-source contributor - \url{https://v1-5-branch.kubeflow.org/docs/about/contributing/}

\Urlmuskip=0mu plus 1mu\relax
\bibliographystyle{unsrt}
\bibliography{bibfile}

\end{document}